\documentclass[12pt]{article}
\usepackage[dvipsnames, table]{xcolor}

  \usepackage[pdftex]{graphicx}

\usepackage{epstopdf}

\usepackage[cmex10]{amsmath}
\usepackage{amsfonts, amssymb, amsthm, mathtools}
\usepackage{empheq}

\usepackage[caption=false, format=hang]{subfig}

\usepackage{multirow}   		
\usepackage{booktabs}   		
\usepackage{array}
\newcolumntype{+}{>{\global\let\currentrowstyle\relax}}
\newcolumntype{^}{>{\currentrowstyle}}
\newcommand{\rowstyle}[1]{\gdef\currentrowstyle{#1}#1\ignorespaces}
\newcommand{\brow}{\rowstyle{\bfseries}}

\usepackage[sort,compress]{cite}
\usepackage{microtype}
\usepackage{algorithm}

\newcommand{\NN}{\ensuremath{\mathbb N}}

\newcommand{\RR}{\ensuremath{\mathbb R}}

\newcommand{\HH}{\ensuremath{{\mathcal H}}}
\newcommand{\GG}{\ensuremath{{\mathcal G}}}

\newcommand{\RX}{\ensuremath{\left]-\infty,+\infty\right]}}

\DeclarePairedDelimiter{\parens}{(}{)}
\DeclarePairedDelimiter{\bracks}{[}{]}

\DeclarePairedDelimiter{\norm}{\lVert}{\rVert}

\newcommand{\minimize}[2]{\ensuremath{\underset{\substack{{#1}}}{\operatorname{minimize}}\;\;#2}}
\newcommand{\scal}[2]{{\left\langle{{#1}\mid{#2}}\right\rangle}}
\newcommand{\menge}[2]{\big\{{#1}~\big |~{#2}\big\}} 

\newcommand{\lev}[1]{{\ensuremath{{{{\operatorname{lev}}}_{\leq #1}}\,}}}
\newcommand{\pinf}{\ensuremath{{+\infty}}}
\DeclareMathOperator{\subto}{s.t.}
\newcommand{\bell}{\boldsymbol{\ell}}
\newcommand{\epi}{\operatorname{epi}}
\newcommand{\Id}{\ensuremath{\operatorname{Id}}}
\newcommand{\dom}{\ensuremath{\operatorname{dom}}}
\newcommand{\prox}{\ensuremath{\operatorname{prox}}}

\newtheorem{theorem}{Theorem}[section]

\newtheorem{proposition}[theorem]{Proposition}

\theoremstyle{definition}

\newcommand{\Diag}{\ensuremath{\operatorname{Diag}}}
\newcommand{\Nx}{K}

\title{A Non-Local Structure Tensor Based Approach for Multicomponent Image Recovery Problems}

\makeatletter
\let\Title\@title
\makeatother

\author{
Giovanni~Chierchia\thanks{Institut Mines-T\'el\'ecom, T\'el\'ecom ParisTech, CNRS LTCI, 75014 Paris, France} \and
Nelly~Pustelnik\thanks{ENS Lyon, Laboratoire de Physique, UMR CNRS 5672, F69007 Lyon, France} \and
B\'eatrice~Pesquet-Popescu\footnotemark[1] \and
and~Jean-Christophe~Pesquet\thanks{Universit\'e Paris-Est, LIGM, UMR CNRS 8049, 77454 Marne-la-Vall\'ee, France}
}

\begin{document}

\maketitle

\begin{abstract}
Non-local total variation (NLTV) has emerged as a useful tool in variational methods for image recovery problems. In this paper, we extend the NLTV-based regularization to multicomponent images by taking advantage of the structure tensor (ST) resulting from the gradient of a multicomponent image. The proposed approach allows us to penalize the non-local variations, jointly for the different components, through various $\bell_{1,p}$-matrix-norms with $p \ge 1$. To facilitate the choice of the hyper-parameters, we adopt a constrained convex optimization approach in which we minimize the data fidelity term subject to a constraint involving the ST-NLTV regularization. The resulting convex optimization problem is solved with a novel epigraphical projection method. This formulation can be efficiently implemented thanks to the flexibility offered by recent primal-dual proximal algorithms. Experiments are carried out for color, multispectral and hyperspectral images. The results demonstrate the interest of introducing a non-local structure tensor regularization and show that the proposed approach leads to significant improvements in terms of convergence speed over current state-of-the-art methods, such as the Alternating Direction Method of Multipliers.
\end{abstract}


\section{Introduction}
Multicomponent images consist of several spatial maps acquired simultaneously from a scene. Well-known examples are color images, which are composed of red, green, and blue components, or spectral images, which divide the electromagnetic spectrum into many components that represent the light intensity across a number of wavelengths. Multicomponent images are often degraded by blur and noise arising from sensor imprecisions or physical limitations, such as aperture effects, motion or atmospheric phenomena. Additionally, a decimation modelled by a sparse or random matrix can be encountered in several applications, such as compressive sensing \cite{Rauhut_2008_j-tit_compres_sensing_dict, Romberg2009_siims_random_convol, Golbabaee_2012_ICASSP_DHSI_CS_LRPS}, inpainting \cite{Fadili2009,Lorenzi2013,Shen2014}, or super-resolution~\cite{Zhang_H_2012_sigproc_super_resol_HSI}. As a consequence, the standard imaging model consists of a blurring operator \cite{Oliveira_2013_j-tip_param_blur_estim_blind} followed by a decimation and a (non-necessarily additive) noise. 

The main focus of this paper is multicomponent image recovery from degraded observations, for which it is of paramount importance to exploit the intrinsic correlations along spatial and spectral dimensions. To this end, we adopt a variational approach based on the introduction of a \textit{non-local total variation structure tensor} (ST-NLTV) regularization and we show how to solve it practically with constrained convex optimization techniques. In addition to more detailed theoretical developments, this paper extends our preliminary work in \cite{Chierchia_2013_epi_approach_NLST} by considering a regularization based on nuclear, Frobenius and spectral norms, by providing a performance evaluation w.r.t.\ two state-of-the-art methods in imaging spectroscopy \cite{Yuan_2012_j-ieee-tgrs_hyper_denoising_TV,Cheng_2014_j-tgrs_inpaint_images_MNLTV}, and by presenting a comparison in terms of execution times with a solution based on the Alternating Direction Method of Multipliers method.

\subsection{ST-NLTV regularization}
The quality of the results obtained through a variational approach strongly depends on the ability to model the regularity present in images. Since natural images are often piecewise smooth, popular regularization models tend to penalize the image gradient. In this context, \textit{total variation} (TV) \cite{Rudin_L_1992_tv_atvmaopiip,Combettes_PL_2004_tip_TV_irstavc} has emerged as a simple, yet successful, convex optimization tool. However, TV fails to preserve textures, details and fine structures, because they are hardly distinguishable from noise. To improve this behaviour, the TV model has been extended by using some generalizations based on higher-order spatial differences \cite{Bredies_2010_siims_TGV,Ono_2014_icassp_TGV_constraint}, higher-degree directional derivatives \cite{Hu_2012_ieee-tip_HDTV,Hu_2014_ieee-tip_generalized_HDTV}, or the non-locality principle \cite{Gilboa_G_2009_j-siam-mms_nonlocal_oai, Sutour_2014_ieee-tip_adap_regul_nlm, Yang_2013_ieee-tip_nl_regul_inv_prob}. Another approach to overcome these limitations is to replace the gradient with an operator that yields a more suitable sparse representation of the image, such as a frame \cite{Mallat_S_1999_ap_wavelet_awtosp, Benazza-Benyahia_2005_j-tip_robust_wavelet_multicomp, Chaux_2008_j-tsp_nonlinear_stein_multichannel, Chaux_C_2010_book_wavelet_tdmi, Briceno_L_2011_j-math-imaging-vis_pro_ami, Foi_2007_j-tip_pointwise_sa_dct} or a learned dictionary \cite{Aharon2006_j-ieee-tsp_K-SVD,Elad2010,Li2014}. In this context, the family of Block Matching 3-D algorithms \cite{Dabov_2007_j-tip_image_den_bm3d,Maggioni_2013_j-tip_volume_den_bm4d} has been recently formulated in terms of analysis and synthesis frames \cite{Danielyan_2012_j-tip_variational_bm3d}, substantiating the use of the non-locality principle as a valuable image modeling tool. 

The extension of TV-based models to multicomponent images is, in general, non trivial. A first approach consists of computing TV channel-by-channel and then summing up the resulting smoothness measures \cite{Blomgren_P_1998_color_tv, Attouche_H_2006_siam_variational_asb, Zach_C_2007_duality_bartof, Duval_V_2009_projected_gcid}. Since there is no coupling of the components, this approach may potentially lead to component smearing and loss of edges across components. An alternative way is to process the components jointly, so as to better reveal details and features that are not visible in each of the components considered separately. This approach naturally arises when the gradient of a multicomponent image is thought of as a \emph{structure tensor} (ST) \cite{DiZenzo_S_1986_gradient_multimage, Sapiro_G_1996_anisotropic_dmicf, Sochen_N_1997_general_fllv, Weickert_J_1999_coherence_edci, Tschumperle_D_2001_p-scia_constrained_upd, Bresson_X_2008_fastdual, Jung_2011_j-tip_nonloca_ms_color_image, Goldluecke_B_2012_natural_tv_gmt, Lefkimmiatis_2013_ssvm_patch_ST}, i.e.\ a matrix that summarizes the prevailing direction of the gradient. The idea behind ST-based regularization is to penalize the eigenvalues of the structure tensor, in order to smooth in the direction of minimal change \cite{Blomgren_P_1998_color_tv}. A concise review of both frameworks can be found in \cite{Goldluecke_B_2012_natural_tv_gmt}, where an efficient ST-TV regularization was suggested for color imagery.

In order to improve the results obtained in the color and hyperspectral restoration literature based on structure tensor, our first main contribution consists of applying the non-locality principle to ST-TV regularization.

\subsection{Constrained formulation}
Regarding the variational formulation of the data recovery problem, one may prefer to adopt a constrained formulation rather than a regularized one. Indeed, it has been recognized for a long time that incorporating constraints directly on the solution often facilitates the choice of the involved parameters \cite{Youla_DC_1982_tmi_POCS_irbtmopocs, Trussell_H_1984_tassp_feasible_ssp, Combettes_PL_1994_tsp_Inconsistent_sfplssiaps, Kose_K_2012_p-icassp_fil_vmd, Teuber_T_2012_report_minimization_pes, Studer_2014_democrat_repres,Ono_2014_icassp_TGV_constraint}. The constraint bounds may be related to some knowledge on the degradation process, such as the noise statistical properties, for which the expected values are known \cite{Teuber_T_2012_report_minimization_pes}. When no specific information about the noise is available, these bounds can be related to some physical properties of the target signal. For example, a reasonable upper bound on the TV constraint may be available for certain classes of images, since TV constitutes a geometrical attribute that exhibits a limited variance over, e.g., views of similar urban areas in satellite imaging, tomographic reconstructions of similar cross sections, fingerprint images, text images and face images \cite{Combettes_PL_2004_tip_TV_irstavc}.

One of the difficulties of constrained approaches is that a closed form of the projection onto the considered constraint set is not always available. Closed forms are known for convex sets such as $\bell_2$-balls, hypercubes, hyperplanes, or half-spaces \cite{Hiriart_Urruty_1996_book_convex_amaIf}. However, more sophisticated constraints are usually necessary in order to effectively restore multicomponent images. Taking advantage of the flexibility offered by recent proximal algorithms, we propose an epigraphical method allowing us to address a wide class of convex constraints. Our second main contribution is thus to provide an efficient solution based on proximal tools in order to solve convex problems involving matricial $\bell_{1,p}$-ball constraints.

\subsection{Imaging spectroscopy}
Spectral imagery is used in a wide range of applications, such as remote sensing \cite{Valero_2013_j-tip_hsi_repr_proc}, astronomical imaging~\cite{Molina_2001_image_resto_astronomy}, and fluorescence microscopy \cite{Vonesh_2006_color_rev_bioimag}. In these contexts, one typically distinguishes between \textit{multispectral} (MS) and \textit{hyperspectral} (HS) images. In general, HS images are capable to achieve a higher spectral resolution than MS images (at the cost of acquiring a few hundred bands), which results in a better spectral characterization of the objects in the scene. This gave rise to a wide array of applications in remote sensing, such as detection and identification of the ground surface \cite{Bernard_2012_j-tip_spect_spat_classif}, as well as military surveillance and historical manuscript research.

The primary characteristic of hyperspectral images is that an entire spectrum is acquired at each point, which implies a huge correlation among close spectral bands. As a result, there has been an emergence of variational methods to efficiently model the spectral-spatial regularity present in such kind of images. To the best of our knowledge, these methods can be roughly divided into three classes.

A first class of approaches consists of extending the regularity models used in color imagery \cite{Tschumperle_2005_tpami_vector_image_pde,Goldluecke_B_2012_natural_tv_gmt}. To cite a few examples, the work in \cite{Zhang_H_2012_sigproc_super_resol_HSI} proposed a super-resolution method based on a component-by-component TV regularization. To deal with the huge size of HS images, the authors performed the actual super-resolution on a few principal image components (obtained by means of PCA), which are then used to interpolate the secondary components. In \cite{Briceno_L_2011_j-math-imaging-vis_pro_ami}, the problem of MS denoising is dealt with by considering a hybrid regularization that induces each component to be sparse in an orthonormal basis, while promoting similarities between the components by means of a distance function applied on wavelet coefficients. Another kind of spectral adaptivity has been proposed in \cite{Yuan_2012_j-ieee-tgrs_hyper_denoising_TV} for HS restoration. It consists of using the multicomponent TV regularization in \cite{Bresson_X_2008_fastdual} that averages the Frobenius norms of the multicomponent gradients. The same authors have recently proposed in \cite{Cheng_2014_j-tgrs_inpaint_images_MNLTV} an inpainting method based on the multicomponent NLTV regularization. The link between this method and the proposed work will be discussed later.

A second class of approaches consists of modeling HS images as three-dimensional tensors, i.e.\ two spatial dimensions and one spectral dimension. First denoising attempts in this direction were pursued in \cite{Martin-Herrero_2007_j-ieee-tgrs_ani_diff_hyper, Mendez-Rial_2010_aniso_hypercube}, where tensor algebra was exploited to jointly analyze the HS hypercube considering vectorial anisotropic diffusion methods. Other strategies, based on filtering, consider tensor denoising methods such as multiway Wiener filtering (see \cite{Lin_2013_eurasip_survey_tensor_decomp} for a survey on this subject).

A third class of approaches is based on robust PCA \cite{Candes_E_2011_j-acm_RPCA} or low-rank and sparse matrix decomposition \cite{Hsu_D_2011_j-ieee-tit_robust_LRPS}. These methods proceed by splitting a HS image into two separate contributions: an image formed by components having similar shapes (low-rank image) and an image that highlights the differences between the components (sparse image). For example, the work in \cite{Golbabaee_2012_ICASSP_DHSI_CS_LRPS} proposed a convex optimization formulation for recovering an HS image from very few compressive-sensing measurements. This approach involved a penalization based on two terms: the $\bell_*$ nuclear norm of the matrix where each column corresponds to the 2D-wavelet coefficients of a spectral band (reshaped in a vector) and the $\bell_{1,2}$-norm of the wavelet-coefficient blocks grouped along the spectral dimension. A similar approach was followed in \cite{Ely_2013_icassp_exploit_struct_HIS}, even though the $\bell_{*}$/$\bell_{1,2}$-norm penalization was applied directly on the HS pixels, rather than using a sparsifying linear transform. 

A third contribution of this work is to adapt the proposed ST-NLTV regularization in order to efficiently deal with reconstruction problems (not only denoising) in the context of imaging spectroscopy. The resulting strategy is based on tensor algebra ideas, but it uses variational strategies rather than anisotropic diffusion or adaptive filtering. Moreover, comparisons with recent works have been performed.

\subsection{Outline}
The paper is organized as follows. Section~\ref{sec:prop_approach} describes the degradation model and formulates the constrained convex optimization problem based on the non-local structure tensor. Section~\ref{sec:algo} explains how to minimize the corresponding objective function via proximal tools. Section~\ref{sec:results} provides an experimental validation in the context of color, MS and HS image restoration. Conclusions are given in Section~\ref{sec:conclusion}.\\

\subsection{Notation}
Let $\HH$ be a real Hilbert space.
$\Gamma_0(\HH)$ denotes the set of proper, lower semicontinuous, convex functions from $\HH$ to $\RX$. Remember that a function 
$\varphi\colon \HH \to \RX$ is proper if its domain $\dom \varphi = \menge{y\in \HH}{\varphi(y)<\pinf}$ is nonempty. 
The subdifferential of $\varphi$ at $x\in \HH$ is $\partial \varphi(x) = \menge{u\in\HH}{(\forall y\in\HH)\;
\scal{y-x}{u}+\varphi(x)\leq \varphi(y)}$.
The epigraph of $\varphi\in \Gamma_0(\HH)$ is the nonempty closed convex subset of $\HH\times \RR$ defined as $\epi \varphi = \menge{(y,\zeta) \in \HH\times \RR}{\varphi(y)\le \zeta}$, the lower level set of $\varphi$ at height $\zeta \in \RR$ is the nonempty closed convex subset of $\HH$ defined as $\lev{\zeta}\varphi = \menge{y\in \HH}{\varphi(y) \le \zeta}$. The projection onto a nonempty closed convex subset $C\subset \HH$ is, for every $ y \in \HH$, 
$P_C(y) = \operatorname{argmin}_{u\in C} \|u-y\|$.
The indicator function $\iota_C$ of $C$ is equal to $0$ on $C$ and $\pinf$ otherwise. Finally, $\Id$ (resp.\ $\Id_N$) denotes the identity operator (resp.\ the identity matrix of size $N \times N$).

\section{Proposed approach}\label{sec:prop_approach}

\subsection{Degradation model}
The $R$-component signal of interest is denoted by $\overline{\textrm x} = (\overline{x}_1,\ldots, \overline{x}_{R})\in (\RR^{N})^R$. In this work, each signal component will generally correspond to an image of size $N = N_1\times N_2$. In imaging spectroscopy, $R$ denotes the number of spectral bands. The degradation model that we consider in this work is
\begin{equation}\label{eq:degrad_model}
{\textrm z} = \mathcal{B}(\textrm A\overline{\textrm x})
\end{equation}
where ${\textrm z}= (z_1,\ldots, z_S)\in (\RR^{\Nx})^S$ denotes the degraded observations, $\mathcal{B}\colon (\RR^{\Nx})^S \to (\RR^{\Nx})^S$ models the effect of a (non-necessarily additive) noise, and $\textrm A= (\textrm A_{s,r})_{1\leq s\leq S,1\leq r \leq R}$ is the degradation linear operator with $\textrm A_{s,r}\in \RR^{\Nx\times N}$, for every $(s,r)\in \{1,\ldots,S\}\times \{1,\ldots,R\}$. In particular, this model can be specialized in some of the applications mentioned in the introduction, such as  super-resolution and compressive sensing, as well as unmixing as explained in the following.
\begin{enumerate}

\item \textbf{Super-resolution} \cite{Zhang_H_2012_sigproc_super_resol_HSI}. In this scenario, ${\textrm z} $ denotes $B$ multicomponent images at low-resolution and $\overline{\textrm x}$ denotes the (high-resolution) multicomponent image to be recovered. The operator $\textrm A$ is a block-diagonal matrix with $S=B R$ while, for every $r \in \{1,\dots,R\}$ and $b \in \{1,\dots, B\}$, 
$\textrm A_{B(r-1)+b,r} = \textrm D_b \textrm T \textrm W_r$ is a composition of a warp matrix $\textrm W_r \in \RR^{N\times N}$, a camera blur operator $\textrm T\in \RR^{N\times N}$, and a downsampling matrix $\textrm D_b \in \RR^{\Nx\times N}$ with $K<N$. The noise is assumed to be a zero-mean  white Gaussian additive noise. It follows that $B$ different noisy decimated versions of the same blurred and
warped component are available.
This yields the following degradation model: for every $r\in\{1,\dots,R\}$ and $b\in\{1,\dots,B\}$,
\begin{equation}
z_{B(r-1)+b} = \textrm D_b \textrm T \textrm W_r\,\overline{x}_r + \varepsilon_{B(r-1)+b} 
\end{equation}
where $\varepsilon_{B(r-1)+b} \sim \mathcal{N}(0,\sigma^2\Id_K)$. 

\vspace{0.5em}
\item \textbf{Compressive sensing} \cite{Golbabaee_2012_ICASSP_DHSI_CS_LRPS}. In this scenario, ${\textrm z} $ denotes the \emph{compressed} multicomponent image and $\overline{\textrm x}$ is the multicomponent image to be reconstructed. The operator $\textrm A$ is a block-diagonal matrix where $S=R$, for every $r \in \{1,\dots,R\}$, $\textrm A_{r,r} = \textrm D_r$, and $\textrm D_r \in \RR^{\Nx\times N}$ is a random measurement matrix with $K\ll N$. The noise is assumed to be a zero-mean white Gaussian additive noise. This leads to the following degradation model:
\begin{equation}\label{eq:cs_model}
(\forall r\in\{1,\dots,R\})\qquad z_r = \textrm D_r\,\overline{x}_r+ \varepsilon_r
\end{equation}
where $\varepsilon_r\sim \mathcal{N}(0,\sigma^2\Id_K)$. 

\vspace{0.5em}
\item \textbf{Unmixing} \cite{Dobigeon_N_j-ieee-tsp_joint_bee,Chouzenoux_E_2013_j-ieee-staeors_fast_cls,Ma_2014_j-tps_sig_proc_hs_unmix}. In this scenario, ${\textrm z} $ models an HS image with $K=N$ having several components whose spectra, denoted by $(\textrm S_r)_{1\leq r \leq R}\in (\RR^{S})^R$, are supposed to be known. The goal is to determine the abundance maps of each component, thus the unknown $\overline{\textrm x}$ models these abundance maps. $R$ denotes the number of components and $S$ the number of spectral measurements. The matrix $\mathrm A$ has a block diagonal structure that leads to the following mixing model: for every $\ell\in \{1,\ldots,N\}$,
\begin{equation}
\begin{bmatrix}
z_1^{(\ell)}\\
\vdots\\
z_S^{(\ell)}
\end{bmatrix}
= \sum_{r=1}^{R} x_r^{(\ell)} \textrm S_r +\varepsilon^{(\ell)},
\end{equation}
where $x_r^{(\ell)}$ is the pixel value for the $r$-th component, $z_s^{(\ell)}$ is the pixel value for the $s$-th spectral measurement, and $\varepsilon^{(\ell)}\sim \mathcal{N}(0,\sigma^2\Id_S)$ denotes the additive noise. 
In this work however, we will not focus  on hyperspectral unmixing that constitutes a specific application area for which tailored algorithms have been developed.
\end{enumerate}

\subsection{Convex optimization problem}
A usual solution to recover $\overline{\textrm x}$ from the observations $\textrm z$ is to follow a convex variational approach that
leads to solving an optimization problem such as
\begin{equation}\label{eq:basic_prob}
\minimize{\textrm x \in C} f(\textrm A \textrm x,\textrm z) \quad \mbox{s.t.} \quad g(\textrm x) \leq\eta,
\end{equation}
where $\eta >0$.  The cost function $f(\cdot,\textrm z) \in \Gamma_0\big((\RR^{\Nx})^S\big)$ aims at insuring that the solution is \emph{close to} the observations. This data term is related to the noise characteristics. For instance, standard choices for
$f$ are a quadratic function for an additive Gaussian noise, an $\bell_1$-norm when a Laplacian noise is involved, and a Kullback-Leibler divergence when dealing with Poisson noise. The function $g \in \Gamma_0\big((\RR^{N})^R\big)$ allows us to impose some regularity on the solution. Some possible choices for this function have been mentioned in the introduction. Finally, $C$ denotes a nonempty closed convex subset of $(\RR^{N})^R$ that can be used to constrain the dynamic range of the target signal, e.g. $C=([0,255]^N)^R$ for standard natural images.

Note that state-of-the-art methods often deal with the regularized version of Problem~\eqref{eq:basic_prob}, that is 
\begin{equation}\label{eq:lagrange_prob}
\minimize{\textrm x \in C} f(\textrm A\textrm x,\textrm z) + \lambda g(\textrm x),
\end{equation}
where $\lambda > 0$. Actually, both formulations are equivalent for some specific values of $\lambda$ and $\eta$. As mentioned in the introduction, the advantage of the constrained formulation is that the choice of $\eta$ may be easier, since it is directly related to the properties of the signal to be recovered.

\subsection{Structure Tensor regularization}
In this work, we propose to model the spatial \emph{and} spectral dependencies in multicomponent images by introducing a regularization 
grounded on the use of a \emph{matrix norm}, which is defined as
\begin{equation}
\big(\forall \textrm x \in (\RR^{N})^R\big)\qquad
g(\textrm x) = \sum_{\ell=1}^N \tau_\ell \Vert \textrm F_\ell \textrm B_\ell \textrm x\Vert_p,
\end{equation}
where $\Vert\cdot\Vert_{p}$ denotes the Schatten ${p}$-norm with ${p}\ge 1$,
$(\tau_\ell)_{1 \le \ell \le N}$ are positive weights, 
and, for every $\ell \in \{1,\dots,N\}$, 
\begin{enumerate}
\item \textbf{block selection}: the operator $\textrm B_\ell \colon (\RR^{N})^R \to \RR^{Q^2\times R}$ selects $Q\times Q$ blocks of each component (including the pixel $\ell$) and rearranges them in a matrix of size $Q^2 \times R$, so leading to
\begin{equation}\label{eq:blk_sel}
\textrm Y^{(\ell)} = 
\begin{bmatrix}
x_1^{(n_{\ell,1})} & \ldots & x_R^{(n_{\ell,1})}\\
\vdots      &        & \vdots\\
x_1^{(n_{\ell,Q^2})} & \ldots & x_R^{(n_{\ell,Q^2})}
\end{bmatrix}
\end{equation}
where $\mathcal{W}_\ell=\{n_{\ell,1},\ldots,n_{\ell,Q^2}\}$ is the set of positions located into the window around $\ell$, and $Q>1$;\footnote{The image borders are handled through symmetric extension.}

\item \textbf{block transform}: the operator $\textrm F_\ell \colon  \RR^{Q^2 \times R} \to \RR^{M_\ell \times R}$ denotes an analysis transform that achieves a sparse representation of grouped blocks, yielding
\begin{equation}\label{eq:ST}
\textrm X^{(\ell)} = \textrm F_\ell \textrm Y^{(\ell)},
\end{equation}
where $M_\ell \le Q^2$.
\end{enumerate}
The resulting structure tensor regularization reads
\begin{equation}\label{eq:mat_norm}
g(\textrm x) = \sum_{\ell = 1}^{N} \tau_\ell\, \Vert \textrm X^{(\ell)} \Vert_{p}.
\end{equation}
Let us denote by 
\begin{equation}
\sigma_{\textrm X^{(\ell)}} = \big(\sigma_{\textrm X^{(\ell)}}^{(m)}\big)_{1\le m\le\widetilde{M}_\ell},\quad \textrm{with } \widetilde{M}_\ell = \min\{M_\ell,R\},
\end{equation}
the singular values of $\textrm X^{(\ell)}$ ordered in decreasing order.
When ${p}\in [1,+\infty[$, we have
\begin{equation}
g(\textrm x) = \sum_{\ell = 1}^{N} \tau_\ell\left(\sum_{m=1}^{\widetilde{M}_\ell} \big(\sigma_{\textrm X^{(\ell)}}^{(m)}\big)^p \right)^{1/p},
\end{equation}
whereas, when ${p}=+\infty$,
\begin{equation}\label{eq:Linf_norm}
g(\textrm x) = \sum_{\ell = 1}^{N} \tau_\ell\,\sigma_{\textrm X^{(\ell)}}^{(1)}.
\end{equation}
When $p=1$, the Schatten norm reduces to the nuclear norm. In such a case, the structure tensor regularization induces a low-rank approximation of matrices $(\textrm X^{(\ell)})_{1\le \ell\le N}$ (see \cite{Stewart_1993_SIAM_history_svd} for a survey on singular value decomposition).

The structure tensor regularization proposed in \eqref{eq:ST} generalizes several state-of-the-art regularization strategies, as explained in the following.

\subsubsection{ST-TV }\label{sec:ST-TV}
We retrieve the multicomponent TV regularization \cite{Bresson_X_2008_fastdual, Goldluecke_B_2012_natural_tv_gmt,Yuan_2012_j-ieee-tgrs_hyper_denoising_TV} by setting $\textrm F_\ell$ to the operator which, for each component index $r \in \{1,\dots,R\}$, computes the difference between $x_r^{(\ell)}$ and its horizontal/vertical nearest neighbours $(x_r^{(\ell_1)},x_r^{(\ell_2)}$), yielding the matrix 
\begin{equation}
\textrm X^{(\ell)}_{_{{\operatorname{TV}}}} = 
\begin{bmatrix}
x_1^{(\ell)} - x_1^{(\ell_1)} & \ldots & x_R^{(\ell)} - x_R^{(\ell_1)}\\
x_1^{(\ell)} - x_1^{(\ell_2)} & \ldots & x_R^{(\ell)} - x_R^{(\ell_2)}\\
\end{bmatrix}
\end{equation}
with $M_\ell = 2$. This implies a $2\times 2$ block selection operator (i.e., $Q=2$). Special cases of ST-TV regularization can be found in the literature when $p=2$ \cite{Yuan_2012_j-ieee-tgrs_hyper_denoising_TV}, or when $F_\ell = \Id$ and $p=1$ \cite{Ono_2013_ieee-tip_convex_regularizer_color}. We will refer to the regularization in \cite{Yuan_2012_j-ieee-tgrs_hyper_denoising_TV} as Hyperspectral-TV in Section~\ref{sec:results}. Moreover, the regularization in \cite{Lefkimmiatis_2013_ssvm_patch_ST} can be seen as an extension of ST-TV arising by setting $\textrm X^{(\ell)} = [\textrm X^{(n)}_{_{{\operatorname{TV}}}}]_{n \in \mathcal{W}_\ell}$, yielding a matrix of size $2\times RQ^2$ (see below \eqref{eq:blk_sel} for the definition of $\mathcal{W}_\ell$). Finally, note that the regularization used in \cite{Zhang_H_2012_sigproc_super_resol_HSI} is intrinsically different from ST-TV, as the former amounts to summing up the smoothed TV \cite{Aujol_JF_2009_jmiv_firstorder_atvbir} evaluated separately over each component.

\subsubsection{ST-NLTV}\label{sec:ST-NLTV}
We extend the NLTV regularization \cite{Gilboa_G_2009_j-siam-mms_nonlocal_oai} to multicomponent images by setting $\textrm F_\ell$ to the operator which, for each component index $r \in \{1,\dots,R\}$, computes the weighted difference between $x_r^{(\ell)}$ and some other pixel values. This results in the matrix 
\begin{equation}\label{e:mattens}
\textrm X^{(\ell)}_{_{{\operatorname{NLTV}}}} = \big[\omega_{\ell,n}(x_r^{(\ell)} - x_r^{(n)})\big]_{n \in \mathcal{N}_\ell, 1 \le r \le R},
\end{equation}
where $\mathcal{N}_\ell \subset \mathcal{W}_\ell\setminus\{\ell\}$ denotes the non-local support of the neighbourhood of $\ell$. Here, $M_\ell$ corresponds to the size of this support. Note that the regularization in \cite{Cheng_2014_j-tgrs_inpaint_images_MNLTV} appears as a special case of the proposed ST-NLTV arising when $p=2$ and the local window is fully used  ($M_\ell = Q^2$). We will refer to it as Multichannel-NLTV in Section~\ref{sec:results}.

For every $\ell \in \{1,\ldots,N\}$ and $n\in \mathcal{N}_\ell$, the weight $\omega_{\ell, n} >0$ depends on the similarity between patches built around the pixels $\ell$ and $n$ of the image to be recovered. Since the degradation process in \eqref{eq:degrad_model} may involve some missing data, we follow a two-step approach in order to estimate these weights. In the first step, the ST-TV regularization is used in order to obtain an estimate $\widetilde{\textrm x}$ of the target image. This estimate  subsequently 
serves in the second step to compute the weights through a \textit{self-similarity} measure as follows:
\begin{equation}\label{eq:weight}
\omega_{\ell,n} = \widetilde{\omega}_\ell\exp\left( - \delta^{-2} \; \norm{\textrm L_\ell \widetilde{\textrm x} - \textrm L_n \widetilde{\textrm x}}^2_2\right),
\end{equation}
where $\delta > 0$, $\textrm L_\ell$ (resp. $\textrm L_n$) selects a $\widetilde{Q} \times \widetilde{Q} \times R$ patch centered at position $\ell$ (resp.\ $n$) after a linear processing depending on the position $\ell$ (resp. $n$), and the constant $\widetilde{\omega}_\ell > 0$ is set so as to normalize the weights (i.e. $\sum_{n \in \mathcal{N}_\ell} \omega_{\ell,n} = 1$). Note that the linear processing is applied to improve the reliability of the self-similarity measure, and thus to insure better image recovery performance. In the simplest case, it consists of point-wise multiplying the selected patches by a bivariate Gaussian function \cite{Buades_A_2005_j-siam-mms_review_idawno}. A more sophisticated processing may involve a convolution with a set of low-pass Gaussian filters whose variances increase as the spatial distance from the patch center grow \cite{Foi_A_2012_p-spie_foveated_ssnif}. For every $\ell \in \{1,\ldots,N\}$, the neighbourhood $\mathcal{N}_\ell$ is built according to the procedure described in \cite{Gilboa_G_2007_j-siam-mms_nonlocal_irss}. In practice, we limit the size of the neighbourhood, so that $M^{(\ell)} \le \overline{M}$ (the choices of $Q$, $\widetilde{Q}$, $\delta$ and $\overline{M}$ are indicated in Section \ref{sec:results}).

\section{Optimization method}\label{sec:algo}
Within the proposed constrained optimization framework, Problem~\eqref{eq:basic_prob} can be reformulated as follows:
\begin{equation}\label{e:prob2}
\minimize{\textrm x \in C} f(\textrm A \textrm x,\textrm z) \quad\subto\quad \Phi \, \textrm x \in D,
\end{equation}
where $\Phi$ is the linear operator defined as
\begin{equation}
\Phi \colon \textrm x \mapsto \textrm X = \left[
\begin{aligned}
&\textrm F_1 \textrm B_1 \textrm x \vphantom{X^{(1)}}\\
&\quad\vdots\\
&\textrm F_N \textrm B_N \textrm x \vphantom{X^{(N)}}
\end{aligned}
\right]
\begin{aligned}
&\vphantom{\textrm F_1 \textrm B_1}\} \; \textrm X^{(1)}\\
&\vphantom{\vdots}\\
&\vphantom{\textrm F_1 \textrm B_1}\} \; \textrm X^{(N)}
\end{aligned}
\end{equation}
with $\textrm X \in \RR^{M \times R}$ and $M = M_1 + \dots + M_N$, while $D$ is the closed convex set defined as
\begin{equation}\label{eq:ST_constraint}
D = \menge{ \textrm X \in \RR^{M\times R}}{\sum_{\ell = 1}^N \tau_\ell \Vert \textrm X^{(\ell)}\Vert_{p} \le \eta }.
\end{equation}
In recent works, iterative procedures were proposed to deal with an $\bell_{1}$- or $\bell_{1,2}$-ball constraint \cite{VanDenBerg_E_2008_j-siam-sci-comp_pro_pfb} and an $\bell_{1,\infty}$-ball constraint \cite{Quattoni_A_2009_p-icml_efficient_plr}. Similar techniques can be used to compute the projection onto $D$, but a more efficient approach consists of adapting the epigraphical splitting technique investigated in \cite{Chierchia_G_2012_j-ieee-tsp_epigraphical_ppt, Harizanov_2013_ssvm_epi_proj_anscombe, Ono_2014_icassp_TGV_constraint, Tofighi_2014_preprint_sig_recon_pesc}.

\subsection{Epigraphical splitting}\label{sec:epi_split}
Epigraphical splitting applies to a convex set that can be expressed as the lower level set of a separable convex function, such as the constraint set $D$ defined in \eqref{eq:ST_constraint}. Some auxiliary variables are introduced into the minimization problem, so that the constraint $D$ can be \emph{equivalently} re-expressed as the intersection of two convex sets. More specifically, different splitting solutions need to be proposed according to the involved Schatten $p$-norm:
\begin{enumerate}
\item in the case when $p=1$, since 
\begin{equation}\label{e:lev_1}
\textrm X \in D \quad\Leftrightarrow\quad \sum_{\ell = 1}^{N}\sum_{m = 1}^{\widetilde{M}_\ell} \tau_\ell\left|\sigma_{\textrm X^{(\ell)}}^{(m)}\right| \le \eta,
\end{equation}
we propose to introduce an auxiliary vector $\zeta \in\RR^{\widetilde{M}}$, with {\small$\zeta = (\zeta^{(\ell,m)})_{1\le\ell\le N,1\le m\le \widetilde{M}_\ell}$} and $\widetilde{M} = \widetilde{M}_1 + \ldots +\widetilde{M}_N$, in order to rewrite \eqref{e:lev_1} as
\begin{equation}
\begin{cases}
(\forall \ell \in \{1,\ldots,N\})(\forall m \in \{1,\ldots,\widetilde{M}_\ell\}) \quad  \left|\sigma_{\textrm X^{(\ell)}}^{(m)}\right| \le \zeta^{(\ell,m)},\\
\displaystyle \; \sum_{\ell=1}^{N}\sum_{m=1}^{\widetilde{M}_\ell} \tau_\ell \, \zeta^{(\ell,m)} \le \eta.
\end{cases}
\end{equation}
Consequently, Constraint \eqref{e:lev_1} is decomposed in two convex sets: a union of epigraphs
\begin{equation}
E = \big\{ {(\textrm X,\zeta) \in \RR^{M\times R} \; \times\; \RR^{\widetilde{M}}}~\big|~(\forall \ell \in \{1,\dots,N\})(\forall m \in \{1,\ldots,\widetilde{M}_\ell\})\quad
(\sigma_{\textrm X^{(\ell)}}^{(m)},\zeta^{(\ell,m)})\in \epi |\cdot| \,\big\},
\end{equation}
and the closed half-space 
\begin{equation}\label{eq:half_space_1}
W = \menge{\zeta\in \RR^{\widetilde{M}}}{\sum_{\ell=1}^{N}\sum_{m=1}^{\widetilde{M}_\ell} \tau_\ell \, \zeta^{(\ell,m)} \le \eta}.
\end{equation}

\item in the case when $p>1$, since
\begin{equation}\label{e:lev_p}
\textrm X \in D \quad\Leftrightarrow\quad \sum_{\ell = 1}^{N}\tau_\ell\left\|\sigma_{\textrm X^{(\ell)}}\right\|_p \le \eta,
\end{equation}
we define an auxiliary vector $\zeta = (\zeta^{(\ell)})_{1\le\ell\le N} \in \RR^N$ of \textit{smaller} dimension $N$, and we rewrite Constraint \eqref{e:lev_p} as
\begin{align}
\begin{cases}
(\forall \ell \in \{1,\ldots,N\})\quad \left\|\sigma_{\textrm X^{(\ell)}}\right\|_p \le \zeta^{(\ell)},\\
\displaystyle \sum_{\ell=1}^{N} \tau_\ell \, \zeta^{(\ell)} \le \eta.
\end{cases}
\end{align}
Similarly to the previous case, Constraint \eqref{e:lev_p} is decomposed in two convex sets: a union of epigraphs
\begin{equation}\label{eq:epi_proj_p}
E = \big\{ {(\textrm X,\zeta) \in \RR^{M\times R} \; \times\; \RR^N}~\big|~
{(\forall \ell \in \{1,\dots,N\})}\quad
(\sigma_{\textrm X^{(\ell)}},\zeta^{(\ell)})\in \epi \Vert \cdot\Vert_{p}\big\},
\end{equation}
and the closed half-space 
\begin{equation}\label{eq:half_space_p}
W = \menge{\zeta\in \RR^{N}}{\sum_{\ell=1}^{N} \tau_\ell \, \zeta^{(\ell)} \le \eta}.
\end{equation}
\end{enumerate}

\subsection{Epigraphical projection}
The epigraphical splitting technique allows us to reformulate Problem~\eqref{e:prob2} in a more tractable way, as follows
\begin{equation}\label{e:prob_epi}
\minimize{(\textrm x,\zeta) \in C \times W} f(\textrm A\textrm x,\textrm z)
\quad\subto\quad
(\Phi \, \textrm x, \, \zeta) \in E.
\end{equation}
The advantage of such a decomposition is that the projections $P_E$ and $P_W$ onto $E$ and $W$ may have closed-form expressions. Indeed, the projection $P_W$ is well-known \cite{Rockafellar_RT_2004_book_Variational_a}, while properties of the projection $P_E$ are summarized in the following proposition, which is straightforwardly proved.

\begin{proposition}\label{ex:epi_proj}
For every $\ell \in \{1,\ldots,N\}$, let 
\begin{equation} 
\textup X^{(\ell)} = \textup U^{(\ell)} \textup S^{(\ell)} \big(\textup V^{(\ell)}\big)^\top
\end{equation}
be the Singular Value Decomposition of $\textup X^{(\ell)} \in \RR^{M_\ell \times R}$, where 
\begin{itemize}
\item $(\textup U^{(\ell)})^\top \textup U^{(\ell)} = \Id_{\widetilde{M}_\ell}$, 
\item $\big(\textup V^{(\ell)}\big)^\top \textup V^{(\ell)} = \Id_{\widetilde{M}_\ell}$,
\item $\textup S^{(\ell)} = \Diag({\sf s}^{(\ell)})$, with ${\sf s}^{(\ell)} =  (\sigma_{\textup X^{(\ell)}}^{(m)})_{1 \le m \le \widetilde{M}_\ell}$.
\end{itemize}
Then,
\begin{equation}
P_E(\textup X,\zeta) = \left(\textup U^{(\ell)} \textup T^{(\ell)} \big(\textup V^{(\ell)}\big)^\top, \; \theta^{(\ell)}\right)_{1\le\ell\le N},
\end{equation}
where $\textup T^{(\ell)} =  \Diag({{\sf t}^{(\ell)}})$ and,
\begin{enumerate}
\item in the case $p=1$, for every $m \in \{1,\dots,\widetilde{M}_\ell\}$
\begin{equation}
({\sf t}^{(\ell,m)} , \; \theta^{(\ell,m)}) = P_{\epi |\cdot|}({\sf s}^{(\ell,m)}, \zeta^{(\ell,m)}),
\end{equation}
\item in the case $p>1$,
\begin{equation}
({\sf t}^{(\ell)} , \; \theta^{(\ell)}) = P_{\epi \Vert\cdot\Vert_{p}}({\sf s}^{(\ell)}, \zeta^{(\ell)}).
\end{equation}
\end{enumerate}
\end{proposition}

\noindent The above result states that the projection onto the epigraph of the $\bell_{1,{p}}$ matrix norm can be deduced from the projection onto the epigraph of the $\bell_{1,{p}}$ vector norm. It turns out that closed-form expressions of the latter projection exist when ${p} \in \{1,2,\pinf\}$ \cite{Chierchia_G_2012_j-ieee-tsp_epigraphical_ppt}. 
For example, for every $({\sf s}^{(\ell)},\zeta^{(\ell)})\in \RR^{\widetilde{M}_\ell}\times \RR$,
\begin{equation}
P_{\epi\Vert \cdot \Vert_2}({\sf s}^{(\ell)},\zeta^{(\ell)})=
\begin{cases}
(0,0), & \mbox{\!\!if $\|{\sf s}^{(\ell)}\|_2 < -\zeta^{(\ell)}$},\\
\displaystyle ({\sf s}^{(\ell)},\zeta^{(\ell)}), & \mbox{\!\!if $\|{\sf s}^{(\ell)}\|_2 < \zeta^{(\ell)}$},\\
\displaystyle
\beta^{(\ell)} \, \big({\sf s}^{(\ell)},\|{\sf s}^{(\ell)}\|_2\big),& \mbox{\!\!otherwise,}
\end{cases}
\label{eq:p2}
\end{equation}
where $\displaystyle \beta^{(\ell)} = \frac{1}{2}\Bigg(1+\frac{\zeta^{(\ell)}}{\|{\sf s}^{(\ell)}\|_2}\Bigg)$. Note that the closed-form expression for $p=1$ can be derived from \eqref{eq:p2}. Moreover,
\begin{equation}
P_{\epi\Vert \cdot \Vert_\infty}({\sf s}^{(\ell)},\zeta^{(\ell)})= ({\sf t}^{(\ell)} , \; \theta^{(\ell)}),
\end{equation}
where, for every ${\sf t}^{(\ell)} = ({\sf t}^{(\ell,m)})_{1\leq m\leq {\widetilde{M}_\ell}} \in \RR^{\widetilde{M}_\ell}$, 
\begin{align}\label{e:projscalinffuncbis}
{\sf t}^{(\ell,m)}&= \min\left\{\sigma_{X^{(\ell)}}^{(m)}, \theta^{(\ell)}\right\},\\
\theta^{(\ell)}&=
\frac{\max\Big\{\zeta^{(\ell)}+\sum_{k=\overline{k}_\ell}^{\widetilde{M}_\ell}\nu^{(\ell,k)},0\Big\}}{{\widetilde{M}_\ell} - \overline{k}_\ell +2}.
\end{align}
Hereabove, $(\nu^{(\ell,k)})_{1\le k \le {\widetilde{M}_\ell}}$ is a sequence of real numbers obtained by sorting $(\sigma_{X^{(\ell)}}^{(m)})_{1\le m \le {\widetilde{M}_\ell}}$ in ascending order
(by setting $\nu^{(\ell,0)} = -\infty$ and $\nu^{(\ell,{\widetilde{M}_\ell}+1)} = \pinf$), and $\overline{k}_\ell$ is the unique integer in {\small $\{1,\ldots,{\widetilde{M}_\ell}+1\}$} such that
\begin{equation}\label{e:projscalmaxfuncterbis}
\nu^{(\ell,\overline{k}_\ell-1)}
< \frac{\zeta^{(\ell)}+\sum_{k=\overline{k}_\ell}^{{\widetilde{M}_\ell}} \nu^{(\ell,k)}}{{\widetilde{M}_\ell}- \overline{k}_\ell +2}
\le \nu^{(\ell,\overline{k}_\ell)}
\end{equation}
(with the convention $\sum_{k=\widetilde{M}^{(\ell)}+1}^{\widetilde{M}^{(\ell)}} \cdot = 0$, i.e.\ the sum is equal to zero when the subscript is greater than the superscript).

Note that the computation of the SVD can be avoided when $p=2$, as the Frobenius norm is equal to the $\bell_2$-norm of the vector of all matrix elements.

\subsection{Proposed algorithm}

The solution of \eqref{e:prob_epi} requires an efficient algorithm for dealing with large scale problems involving nonsmooth functions and linear operators that are non-necessarily circulant. For this reason, we resort here to proximal algorithms \cite{Daubechies_I_2004_cpamath_iterative_talipsc, Chaux_C_2007_j-ip_variational_ffbip, Combettes_PL_2007_istsp_Douglas_rsatncvsr, Figueiredo_M_2007_j-ieee-sel-topics-sp_gra_psr, Beck_A_2009_j-siam-is_fast_istalip, Fornasier_M_2009_j-sjna_subspace_cmtvl1m, Steidl_G_2010_j-math-imaging-vis_removing_mndrsm, Combettes_P_2010_inbook_proximal_smsp, Pesquet_J_2012_j-pjpjoo_par_ipo, Chen_G_1994_j-mp_pro_bdm, Esser_E_2010_j-siam-is_gen_fcf, Chambolle_A_2010_first_opdacpai, Briceno_L_2011_j-siam-opt_mon_ssm, Combettes_P_2011_j-svva_pri_dsa, Vu_B_2011_j-acm_spl_adm, Condat_L_2012,Chen_2013_ip_primal_dual_fpa,Komodakis_2014_playing_duality}. The key tool in these methods is the proximity operator~\cite{Moreau_J_1965_bsmf_Proximite_eddueh} of a function $\phi \in \Gamma_0(\HH)$ on a real Hilbert space, defined as
\begin{equation}
(\forall \textrm u \in \HH)\qquad
\prox_\phi(\textrm u)= \underset{\textrm v\in \HH}{\operatorname{argmin}} \frac12 \|\textrm v-\textrm u\|^2 +
\phi(\textrm v).
\end{equation}
The proximity operator can be interpreted as an implicit subgradient step for the function $\phi$, since $\textrm p = \prox_\phi(\textrm u)$ is uniquely defined through the inclusion $\textrm u - \textrm p \in \partial \phi(\textrm p)$. Proximity operators enjoy many interesting properties \cite{Chaux_C_2007_j-ip_variational_ffbip}. In particular, they generalize the notion of projection onto a closed convex set $C$, in the sense that $\prox_{\iota_C} = P_C$. Hence, proximal methods provide a unifying framework that allows one to address a wide class of convex optimization problems involving non-smooth penalizations and hard constraints.

Among the wide array of existing proximal algorithms, we employ the primal-dual M+LFBF algorithm recently proposed in \cite{Combettes_P_2011_j-svva_pri_dsa}, which is able to address general convex optimization problems involving nonsmooth functions and linear operators without requiring any matrix inversion. This algorithm is able to solve numerically:
\begin{equation}\label{e:formMLFBF}
\minimize{{\rm v} \in \HH}{\phi({\rm v}) + \sum_{i=1}^I} \psi_i(\textrm T_i {\rm v}) + \varphi({\rm v}).
\end{equation}
where $\phi\in \Gamma_0(\HH)$, for every $i \in \{1,\dots,I\}$, $\textrm T_i\colon \HH\to\GG_i$ is a bounded linear operator, $\psi_i \in \Gamma_0(\GG_i)$ and $\varphi\colon \HH\to\RX$ is a convex differentiable function with a $\mu$-Lipschitzian gradient. Our minimization problem fits nicely into this framework by setting $\HH = (\RR^{N})^R \times \RR^{L}$ with
\begin{equation}
L = 
\begin{cases}
\widetilde{M}   &\quad \textrm{if } p=1,\\
N &\quad \textrm{if } p>1,
\end{cases}
\end{equation}
and ${\rm v} = ({\textrm x},\zeta)$. Indeed, we set $I=2$, $\GG_1 = \RR^{M \times R} \times \RR^L$ and $\GG_2= (\RR^{K})^S$ in \eqref{e:formMLFBF}, the linear operators reduce to 
\begin{equation}
\textrm T_1 = \begin{bmatrix}\Phi & 0\\ 0 & \Id_L\end{bmatrix},
\qquad
\textrm T_2 = \begin{bmatrix}\textrm A & 0\end{bmatrix},
\end{equation}
and the functions are as follows
\begin{equation}
\begin{aligned}
&(\forall (\textrm x,\zeta)\in (\RR^{N})^R\times \RR^L)     && \phi({\textrm x},\zeta) &&= \iota_C(\textrm x) + \iota_W(\zeta),\\
&(\forall (\textrm X,\zeta)\in \RR^{M \times R}\times \RR^L) 		&& \psi_1(\textrm X,\zeta)     &&= \iota_E(\textrm X,\zeta),\\
&(\forall \textrm y \in (\RR^{K})^S)                         		&& \psi_2(\textrm y)           &&= f(\textrm y,\textrm z),\\
&(\forall (\textrm x,\zeta)\in (\RR^{N})^R\times \RR^L)     && \varphi(\textrm x,\zeta) &&= 0.
\end{aligned}
\end{equation}
The iterations associated with Problem~\eqref{e:prob_epi} are summarized in Algorithm~\ref{algo:epi}, where 
$\textrm A^\top$ and $\Phi^\top$ designate the adjoint operators of $\textrm A$ and $\Phi$. The sequence $({\rm x}^{[t]})_{t\in \NN}$ generated by the algorithm is guaranteed to converge to a solution to \eqref{e:prob_epi} (see \cite{Combettes_P_2011_j-svva_pri_dsa}).

\subsection{Approach based on ADMM}
Note that an alternative approach to deal with Problem~\eqref{e:prob2} consists of employing the Alternating Direction Method of Multipliers (ADMM) \cite{Eckstein_1992_ADMM} or one of its parallel versions \cite{Setzer_S_2009_j-jvcir_deblurring_pibsbt,Afonso_M_2009_j-tip_augmented_lacofiip,Eckstein1994_j_optim_parallel_ADMM,Combettes_P_2010_inbook_proximal_smsp,Combettes_PL_2008_j-ip_proximal_apdmfscvip,Pesquet_J_2012_j-pjpjoo_par_ipo}, sometimes referred to as the Simultaneous Direction Method of Multipliers (SDMM). Although these algorithms are appealing, they require to invert the operator $\Id + \Phi^\top \Phi + \textrm A^\top \textrm A$, which makes them less attractive than primal-dual algorithms for solving Problem~\eqref{e:prob2}. Indeed, this matrix is not diagonalizable in the DFT domain (due to the form of~$\Phi$), which rules out efficient inversion techniques such as those employed in \cite{Afonso_M_2010_j-tip_fast_iruvsco, Afonso_M_2009_j-tip_augmented_lacofiip, Pustelnik_N_2012_ieee-tsp_rel_tfc}. To the best of our knowledge, this issue can be circumvented in specific cases only, for example when $\Phi$ denotes the NLTV operator defined in \eqref{e:mattens}. In this case, one may resort to the solution in \cite{Peyre_G_2011_p-eusipco_gro_sop, Briceno_L_2011_j-math-imaging-vis_pro_ami}, which consists of decomposing $\Phi$ as follows:
\begin{equation}
\Phi = 
\Omega
\underbrace{
\begin{bmatrix}
\textrm G_1\\
\vdots\\
\textrm G_{Q^2-1}
\end{bmatrix}
}_{\displaystyle \textrm G},
\end{equation}
where, for every $q \in \{1,\dots,Q^2-1\}$, $\textrm G_q \colon (\RR^N)^R \to \RR^{N\times R}$ is a discrete difference operator and $\Omega \in \RR^{M\times N(Q^2-1)}$ is a weighted block-selection operator. So doing, Problem~\eqref{e:prob2} can be \emph{equivalently} reformulated by introducing an auxiliary variable $\xi = \Phi \textrm x \in \RR^{M\times R}$, yielding
\begin{equation}\label{eq:prob_sdmm}
\minimize{(\textrm x, \xi) \in C\times D} f(\textrm A \textrm x, \textrm z) \quad\subto\quad (\textrm G\textrm x,\xi) \in V,
\end{equation}
where $V = \menge{(\textrm X, \xi) \in \RR^{N(Q^2-1)\times R} \times \RR^{M\times R}}{\Omega \textrm X = \xi}$. The iterations associated to SDMM are illustrated in Algorithm~\ref{algo:sdmm}. 

It is worth emphasizing that SDMM still requires to compute the projection onto $D$, which may be done by either resorting to specific numerical solutions \cite{VanDenBerg_E_2008_j-siam-sci-comp_pro_pfb, Weiss_P_2009_Efficient_stvm, Quattoni_A_2009_p-icml_efficient_plr, Fadili_J_2011_tip_tv_proj_fos} or employing the epigraphical splitting technique presented in Section~\ref{sec:epi_split}. However, according to our simulations (see Section~\ref{sec:results:sdmm}), both approaches are slower than Algorithm~\ref{algo:epi}.

\begin{algorithm}
\caption{M+LFBF for solving Problem~\eqref{e:prob_epi}}\label{algo:epi}
{\small
\[
\begin{array}{l}
\mathrm{Initialization}\\
\left\lfloor
\begin{array}{l}
\textrm Y_1^{[0]}\in \RR^{M\times R}, \nu_1^{[0]} \in \RR^L\\
\textrm y_2^{[0]} \in (\RR^K)^S\\
\textrm x^{[0]} \in (\RR^{N})^R, \zeta^{[0]}\in \RR^L\\
\theta = \sqrt{\norm{\textrm A}^2 + \max\{\norm{\Phi}^2,1\}}\\
\epsilon \in ]0, \frac{1}{\theta+1}[\\
\end{array}
\right.\\ 
\mathrm{For}\; t = 0, 1, \dots\\
\left\lfloor
\begin{array}{l}
\displaystyle\gamma_t \in \bracks*{\epsilon, \frac{1-\epsilon}{\theta}}\\
\parens[\Big]{\widehat{\textrm x}^{[t]},\widehat{\zeta}^{[t]}} = \parens[\Big]{\textrm x^{[t]},\zeta^{[t]}} - \gamma_t \parens[\Big]{\Phi^\top \textrm Y_1^{[t]} + \textrm A^\top \textrm y_2^{[t]}, \nu^{[t]}}\\
\parens[\Big]{\textrm p^{[t]},\rho^{[t]}} = \parens[\Big]{P_{C}(\widehat{\textrm x}^{[t]}),P_{W}(\widehat{\zeta}^{[t]})}\\
\parens[\Big]{\widehat{\textrm Y}_1^{[t]},\widehat{\nu}_1^{[t]}} = \parens[\Big]{\textrm Y_1^{[t]},\nu_1^{[t]}} + \gamma_t \parens[\Big]{\Phi \textrm x^{[t]} , \zeta^{[t]}}\\
\parens[\Big]{\widetilde{\textrm Y}_1^{[t]},\widetilde{\nu}_1^{[t]}} = \parens[\Big]{\widehat{\textrm Y}_1^{[t]},\widehat{\nu}_1^{[t]}} - \gamma_t P_{E}\parens[\Big]{\widehat{\textrm Y}_1^{[t]}/\gamma_t,\widehat{\nu}_1^{[t]}/\gamma_t}\\
\parens[\Big]{ \textrm Y_1^{[t+1]},\nu_1^{[t+1]}} = \parens[\Big]{\widetilde{\textrm Y}_1^{[t]},\widetilde{\nu}_1^{[t]}} + \gamma_t \parens[\Big]{\Phi (\textrm p^{[t]}-\textrm x^{[t]}), \rho^{[t]}-\zeta^{[t]} }\\
\widehat{\textrm y}_2^{[t]} = \textrm y_2^{[t]} + \gamma_t \textrm A \textrm x^{[t]} \\
\widetilde{\textrm y}_2^{[t]} = \widehat{\textrm y}_2^{[t]} - \gamma_t \prox_{\gamma_t^{-1}f}\parens[\Big]{\widehat{\textrm y}_2^{[t]}/\gamma_t}\\
\textrm y_2^{[t+1]} = \widetilde{\textrm y}_2^{[t]} + \gamma_t \textrm A (\textrm p^{[t]}-\textrm x^{[t]}) \\
\parens[\Big]{\widetilde{\textrm x}^{[t]},\widetilde{\zeta}^{[t]}} = \parens[\Big]{\textrm p^{[t]},\rho^{[t]}} - \gamma_t \parens[\Big]{ \Phi^\top \widetilde{\textrm Y}_1^{[t]} + \textrm A^\top  \widetilde{\textrm y}_2^{[t]},\widetilde{\nu}_1^{[t]}}\\
\parens[\Big]{\textrm x^{[t+1]},\zeta^{[t+1]}} = \parens[\Big]{\textrm x^{[t]}- \widehat{\textrm x}^{[t]}+ \widetilde{\textrm x}^{[t]},\zeta^{[t]} - \widehat{\zeta}^{[t]} + \widetilde{\zeta}^{[t]}}\\
\end{array}
\right.
\end{array}
\]
}
\end{algorithm}

\begin{algorithm}
\caption{SDMM for solving Problem~\eqref{eq:prob_sdmm}}\label{algo:sdmm}
{\small
\[
\begin{array}{l}
\mathrm{Initialization}\\
\left\lfloor
\begin{array}{l}
\textrm y_1^{[0]} \in (\RR^{N})^R, \textrm Y_2^{[0]} \in \RR^{M\times R}, \textrm y_3^{[0]} \in (\RR^{K})^S\\
\overline{\textrm y}_1^{[0]} \in (\RR^{N})^R, \overline{\textrm Y}_2^{[0]} \in \RR^{M\times R}, \overline{\textrm y}_3^{[0]} \in (\RR^{K})^S\\
\chi_1^{[0]} \in \RR^{M\times R}, \chi_2^{[0]}\in \RR^{M\times R}\\
\overline{\chi}_1^{[0]} \in \RR^{M\times R}, \overline{\chi}_2^{[0]}\in \RR^{M\times R}\\
\textrm H = \Id+\textrm G^\top \textrm G+\textrm A^\top \textrm A
\end{array}
\right.\\
\mathrm{For}\; t = 0, 1, \dots\\
\left\lfloor
\begin{array}{l}
\displaystyle\gamma_t \in \left]0, +\infty\right[\\
 \textrm x^{[t]} = \textrm H^{-1}\left[\textrm y_1^{[t]}-   \overline{\textrm y}_1^{[t]} + \textrm G^\top( \textrm Y_2^{[t]}-  \overline{\textrm Y}_2^{[t]})+ \textrm A^\top( \textrm y_3^{[t]}-  \overline{\textrm y}_3^{[t]})\right]\\
\xi^{[t]} = \frac{1}{2} \left({\chi}_1^{[t]}- \overline{\chi}_1^{[t]}\right) + \frac{1}{2} \left( {\chi}_2^{[t]}- \overline{\chi}_2^{[t]}\right)\\
\textrm y_1^{[t+1]} = P_{C}\left( \textrm x^{[t]}+ \overline{\textrm y}_1^{[t]}\right)\\ 
{\chi}_1^{[t+1]} = P_{D}\left( \xi^{[t]}+ \overline{\chi}_1^{[t]}\right)\\
\big(\textrm Y_2^{[t+1]},{\chi}_2^{[t+1]}\big) = P_V(\textrm G \textrm x^{[t]}+ \overline{\textrm Y}_2^{[t]}, \xi^{[t]}+ \overline{\chi}_2^{[t]})\\
\textrm y_3^{[t+1]} = \prox_{\gamma_t f}\left(\textrm A \textrm x^{[t]}+ \overline{\textrm y}_3^{[t]}\right)\\
\overline{\textrm y}^{[t+1]}_1 = \overline{\textrm y}_1^{[t]} + \textrm x^{[t]} - \textrm y_1^{[t+1]}\\
\overline{\chi}^{[t+1]}_1=  \overline{\chi}_1^{[t]} + \xi^{[t]} - \chi_1^{[t+1]}\\
\big(\overline{\textrm Y}^{[t+1]}_2,\overline{\chi}^{[t+1]}_2\big) =  \big( \overline{\textrm Y}_2^{[t]} + \textrm G \textrm x^{[t]} - \textrm Y_2^{[t+1]}, \overline{\chi}_2^{[t]} + \xi^{[t]} - \chi_2^{[t+1]} \big)\\
\overline{\textrm y}^{[t+1]}_3 =  \overline{\textrm y}_3^{[t]} + \textrm A \textrm x^{[t]} -  \textrm y_3^{[t+1]}\\
\end{array}
\right.
\end{array}
\]
}
\end{algorithm}

\section{Numerical results}\label{sec:results}

\subsection{Color photography}
In this section, we numerically evaluate the ST-NLTV regularization proposed in Section~\ref{sec:prop_approach} and compare it with a standard \emph{channel-by-channel} (CC) regularization \cite{Attouche_H_2006_siam_variational_asb,Chierchia_G_2012_j-ieee-tsp_epigraphical_ppt}:
\begin{equation}
g_{\textrm{cc}}(\textrm x) = \sum_{r=1}^R\sum_{\ell=1}^N \tau_\ell \|X_r^{(\ell)}\|_p,
\end{equation}
where $X_r^{(\ell)}$ denotes the $r$-th column vector of matrix $\textrm X^{(\ell)}$ defined in~\eqref{eq:ST}. Note that, for both CC-NLTV and ST-NLTV, we set $\tau_\ell\equiv 1$, $Q = 11$, $\widetilde{Q} = 5$, $\delta = 35$ and $\overline{M} = 14$, as this setting was observed to yield the best numerical results.

The first experiment is focused on color imaging, i.e.\ the case $R = S = 3$. The noisy observations are obtained with the degradation model in \eqref{eq:cs_model}, where the measurement operator $(\textrm D_r)_{1\le r\le R}$ denotes a decimated convolution. While it is common for color imaging to work in a luminance-chrominance space \cite{Condat2012_ICIP_demosaicking_denoising_TV, Condat2014_j-ieee-spt_prox_algo_TV}, or a perceptually-uniform space \cite{Goldluecke_B_2012_natural_tv_gmt}, the random decimation prevents us from following this approach, because pixels having missing colors cannot be correctly projected onto a different color space. The experiments are thus conducted in the RGB color space, and the dynamic range constraint set $C$ imposes that the pixel values belong to $[0, 255]$. Moreover, the fidelity term related to the noise log-likelihood is $f = \|\textrm A\cdot-\textrm z\|^2_2$.

\begin{figure*}
\centering
\subfloat[Original.]{\includegraphics[width=0.3\textwidth]{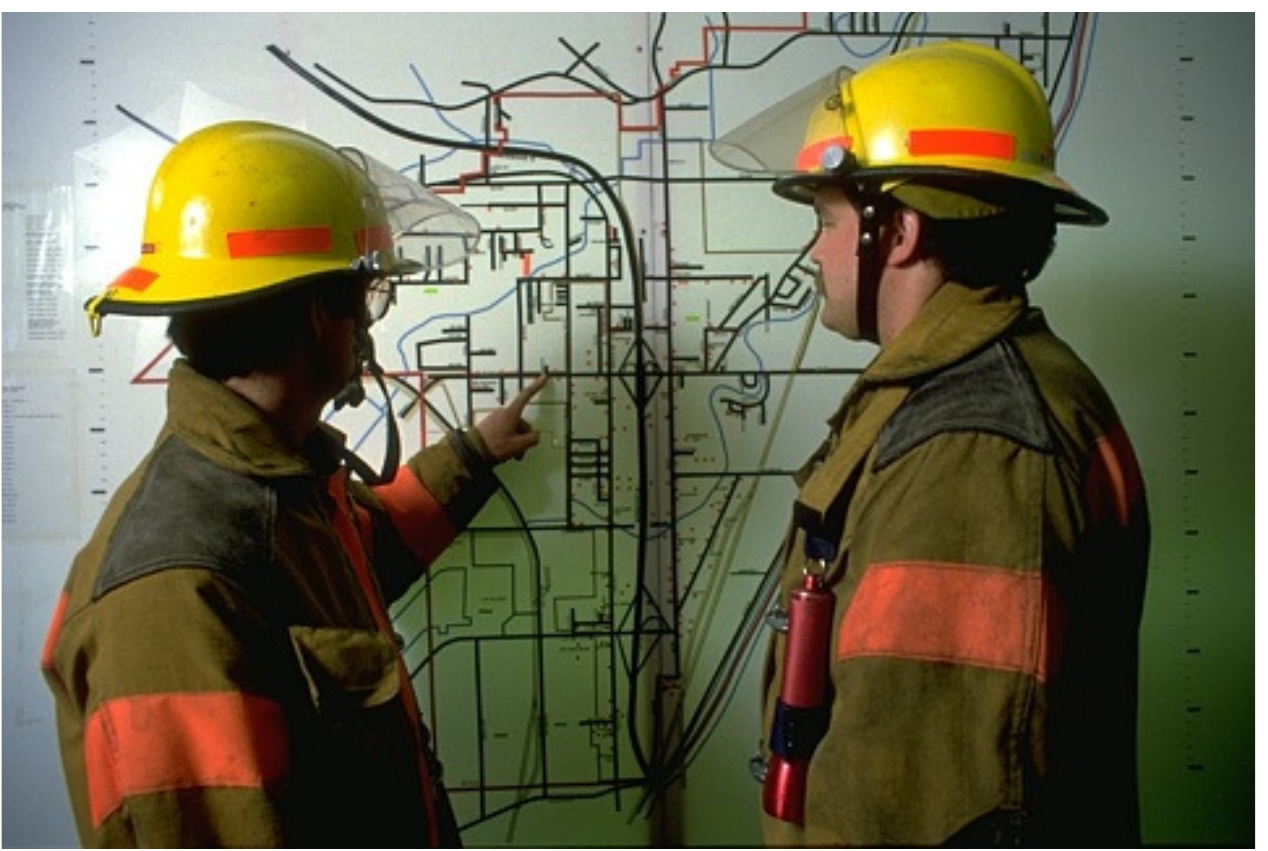}}
\hfill
\subfloat[Noisy.]{\includegraphics[width=0.3\textwidth]{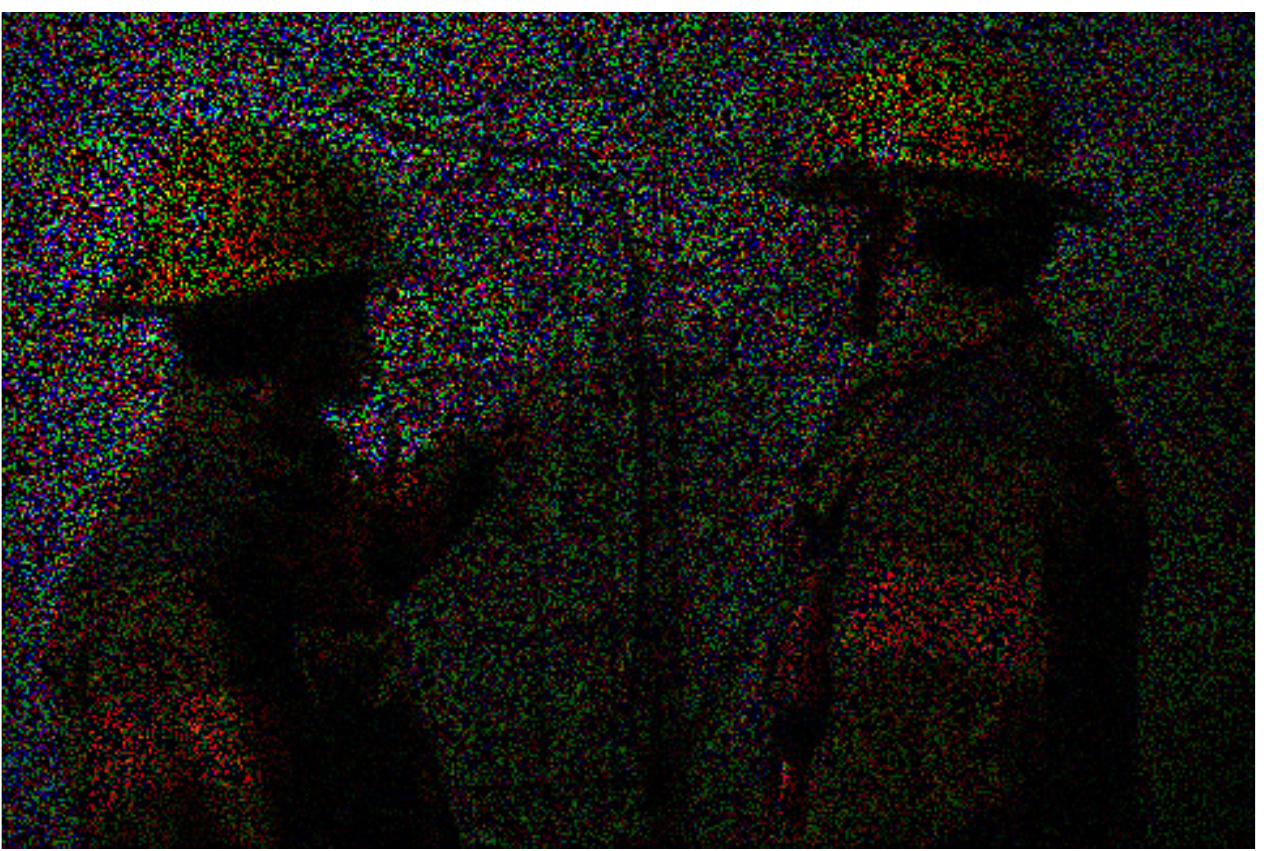}}
\hfill
\subfloat[Zoom.]{\includegraphics[width=0.3\textwidth,crop]{original}}

\subfloat[$\bell_1$-CC-TV \cite{Attouche_H_2006_siam_variational_asb}: 16.15 dB.]{\includegraphics[width=0.3\textwidth,crop]{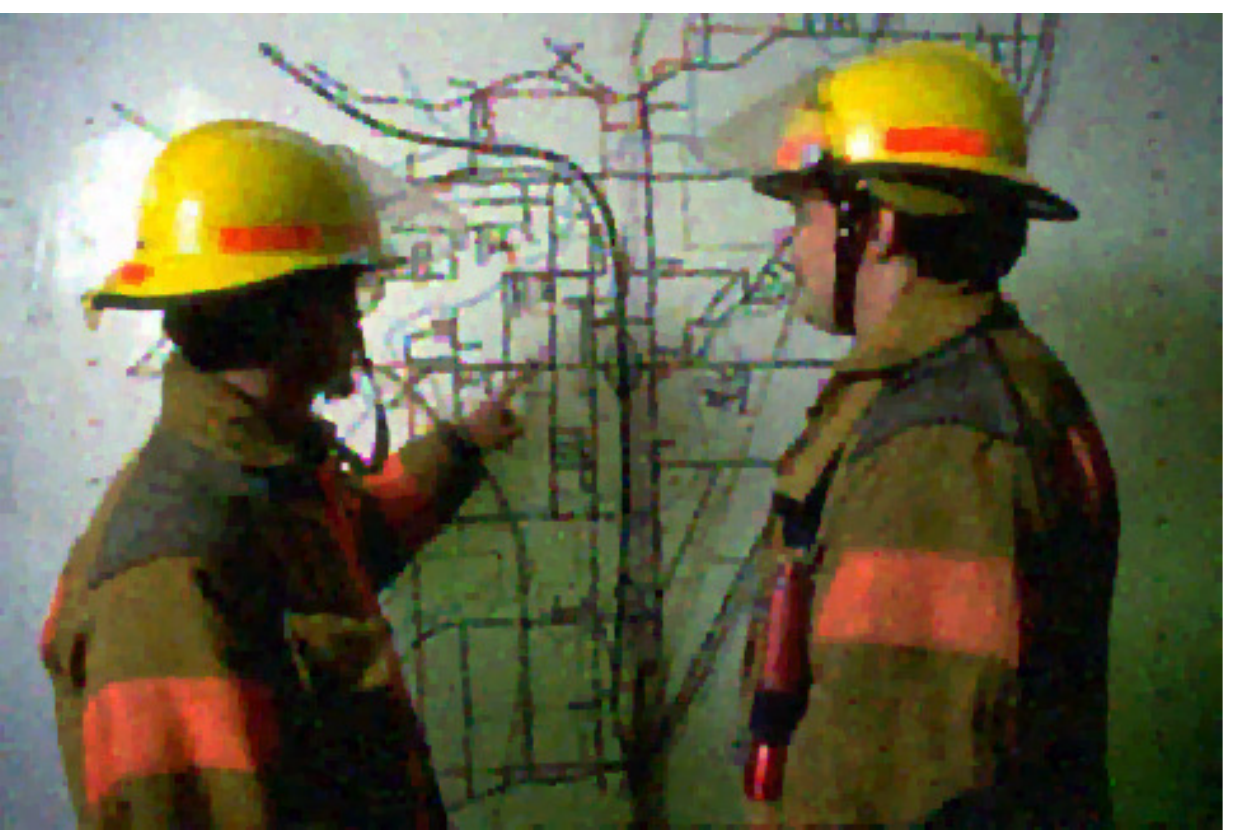}}
\hfill
\subfloat[$\bell_2$-CC-TV \cite{Attouche_H_2006_siam_variational_asb}: 16.32 dB.]{\includegraphics[width=0.3\textwidth,crop]{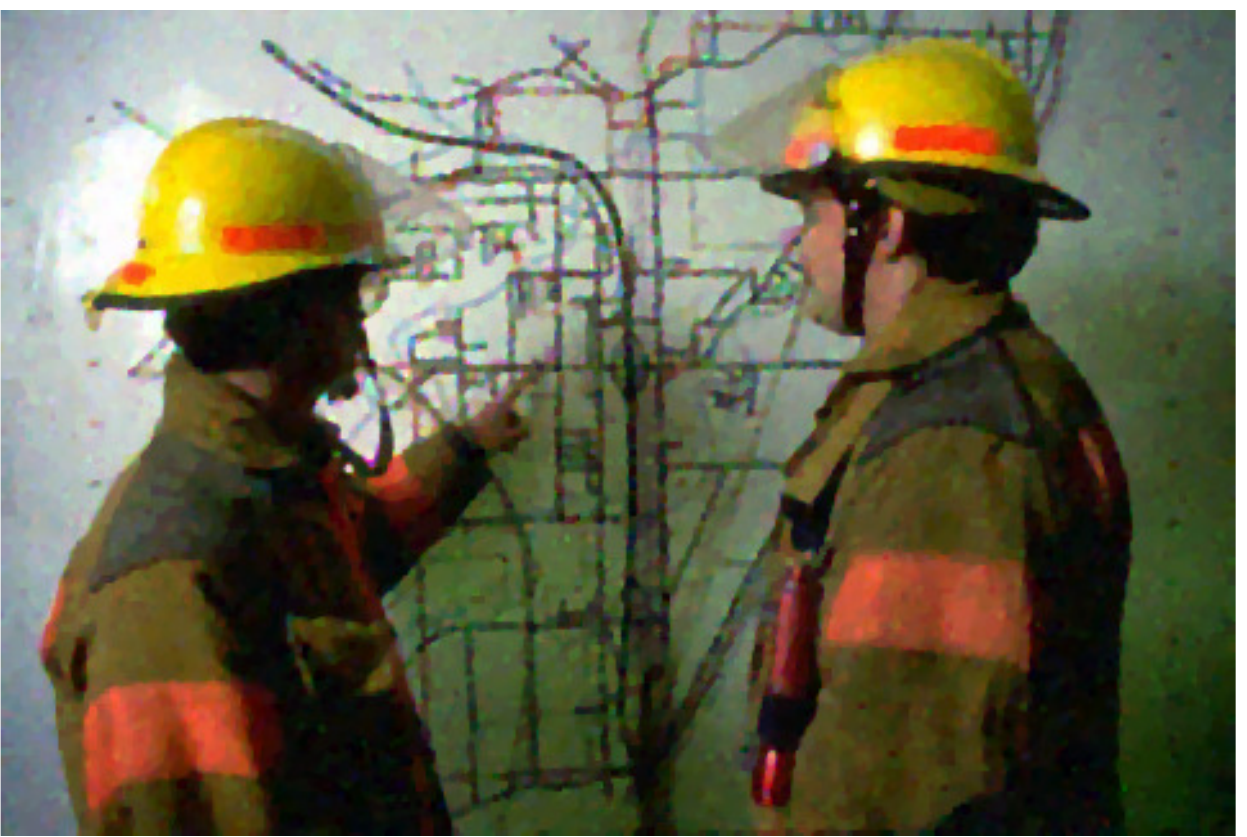}}
\hfill
\subfloat[$\bell_\infty$-CC-TV \cite{Attouche_H_2006_siam_variational_asb}: 16.05 dB.]{\includegraphics[width=0.3\textwidth,crop]{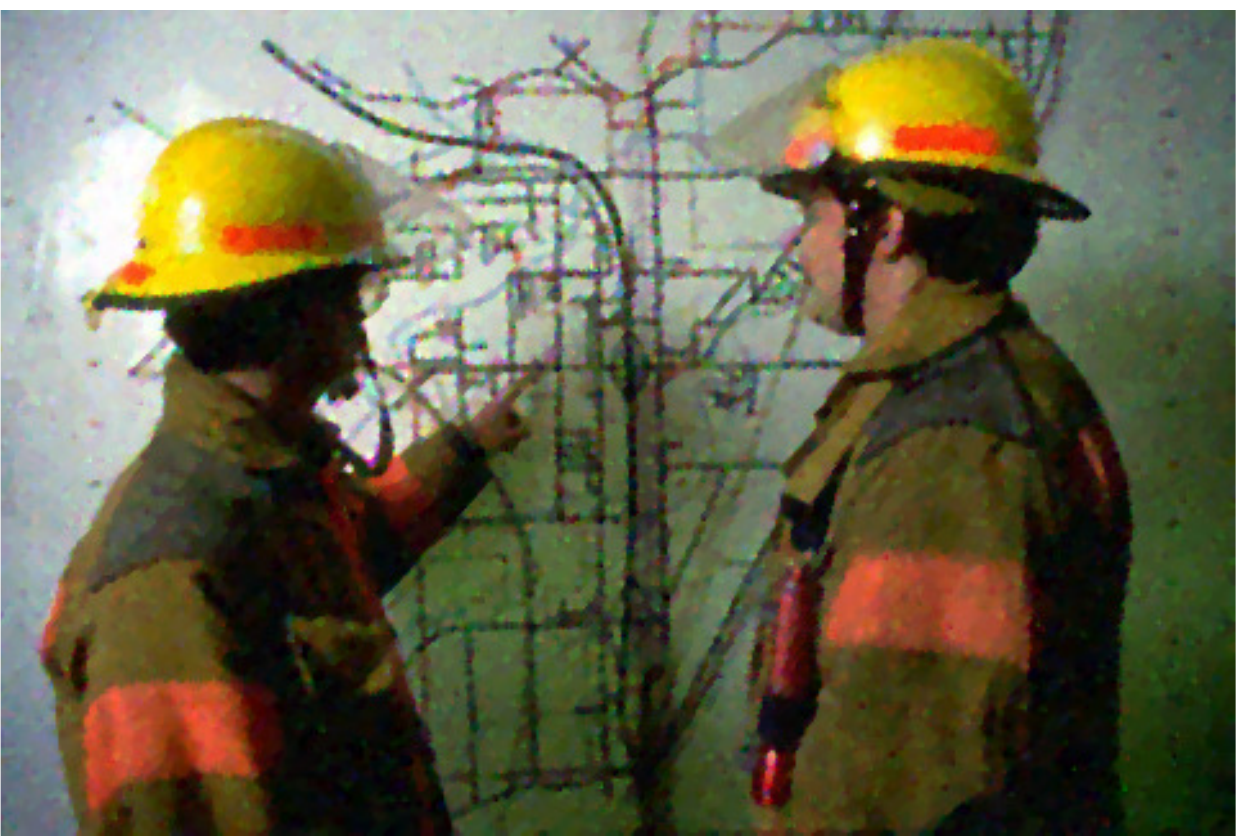}}

\subfloat[$\bell_1$-CC-NLTV \cite{Chierchia_G_2012_j-ieee-tsp_epigraphical_ppt}: 16.87 dB.]{\includegraphics[width=0.3\textwidth,crop]{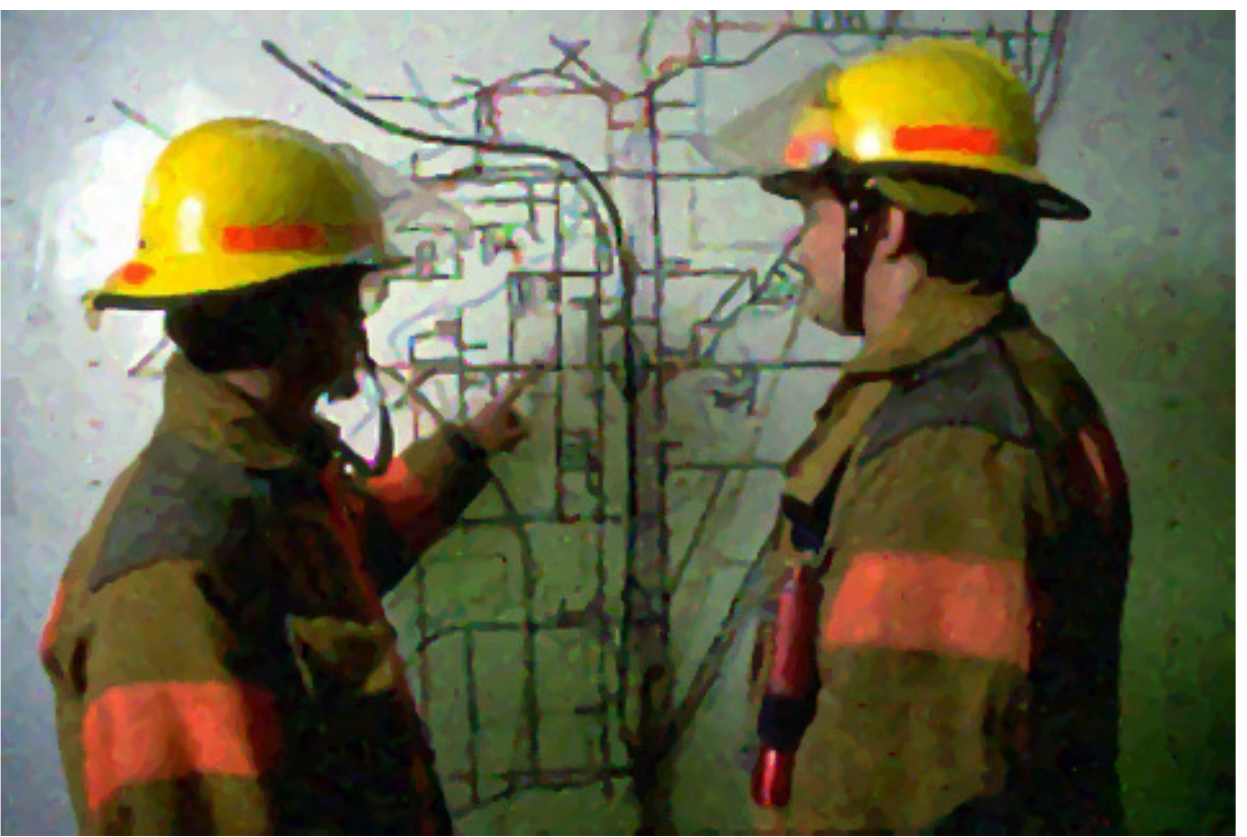}}
\hfill
\subfloat[$\bell_2$-CC-NLTV \cite{Chierchia_G_2012_j-ieee-tsp_epigraphical_ppt}: 17.20 dB.]{\includegraphics[width=0.3\textwidth,crop]{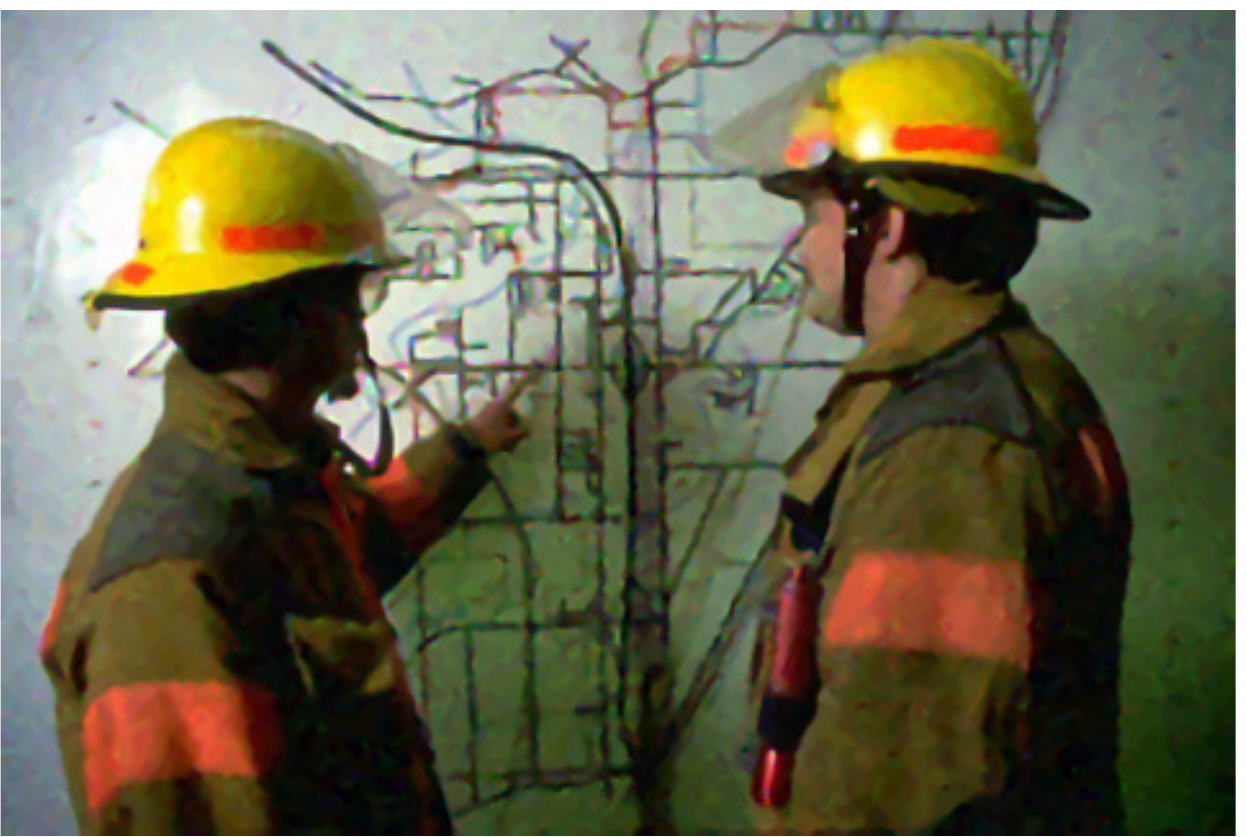}}
\hfill
\subfloat[$\bell_\infty$-CC-NLTV \cite{Chierchia_G_2012_j-ieee-tsp_epigraphical_ppt}: 17.22 dB.]{\includegraphics[width=0.3\textwidth,crop]{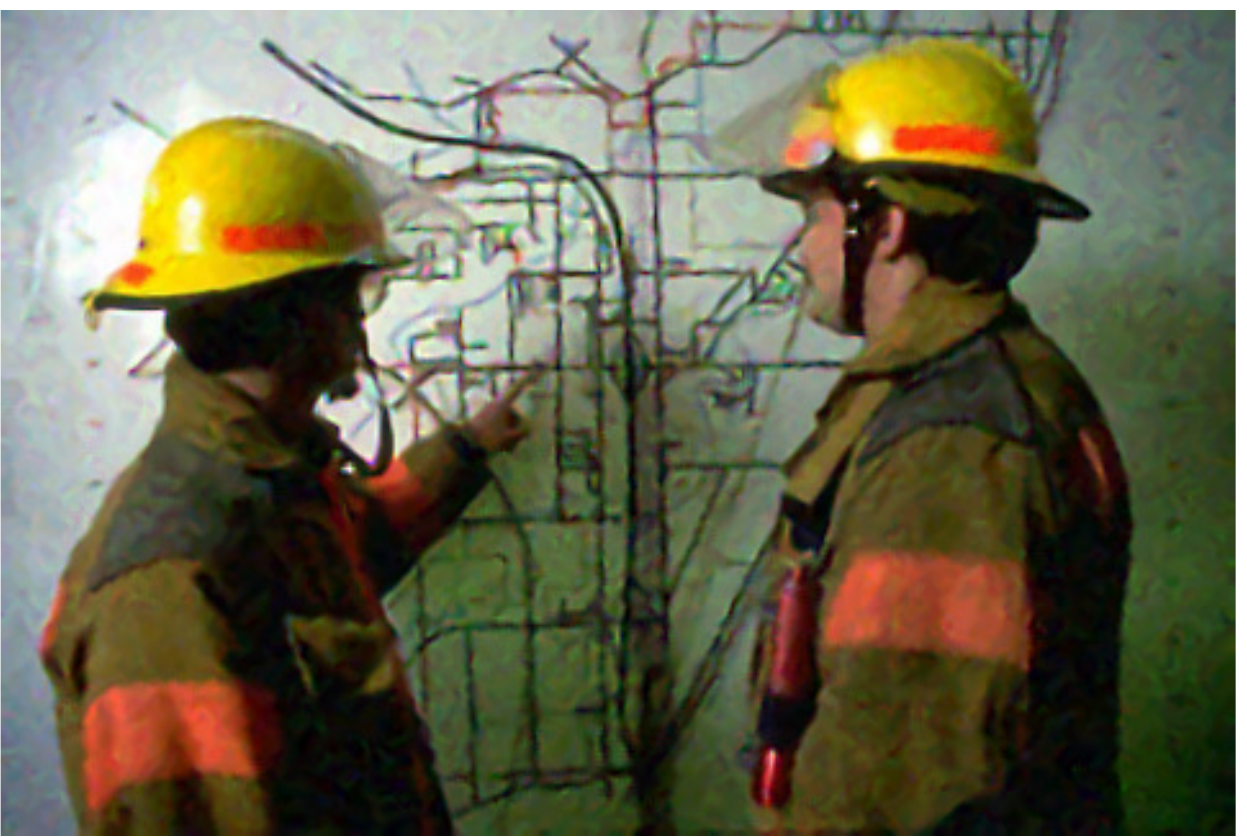}}

\subfloat[$\bell_1$-ST-TV: 17.08 dB.]{\includegraphics[width=0.3\textwidth,crop]{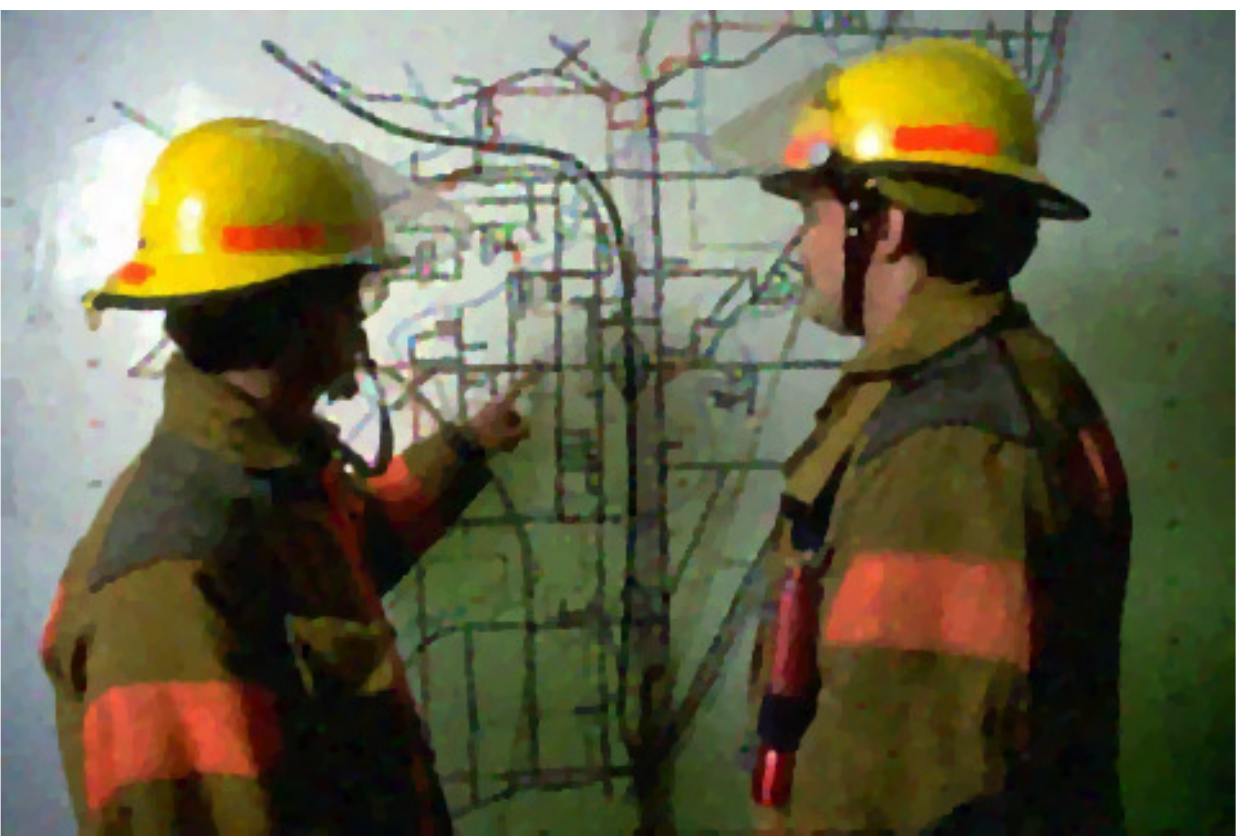}}
\hfill
\subfloat[$\bell_2$-ST-TV \cite{Bresson_X_2008_fastdual}: 16.84 dB.]{\includegraphics[width=0.3\textwidth,crop]{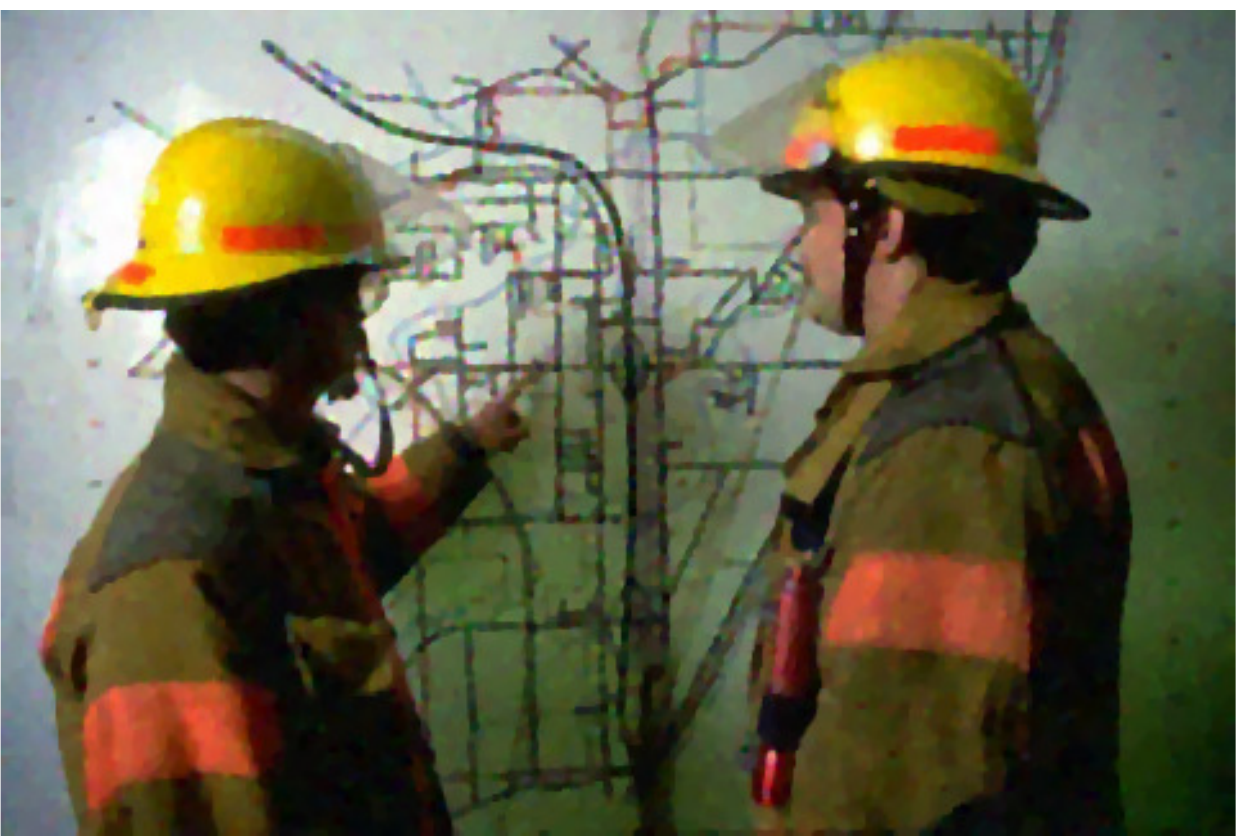}}
\hfill
\subfloat[$\bell_\infty$-ST-TV \cite{Goldluecke_B_2012_natural_tv_gmt}: 16.43 dB.]{\includegraphics[width=0.3\textwidth,crop]{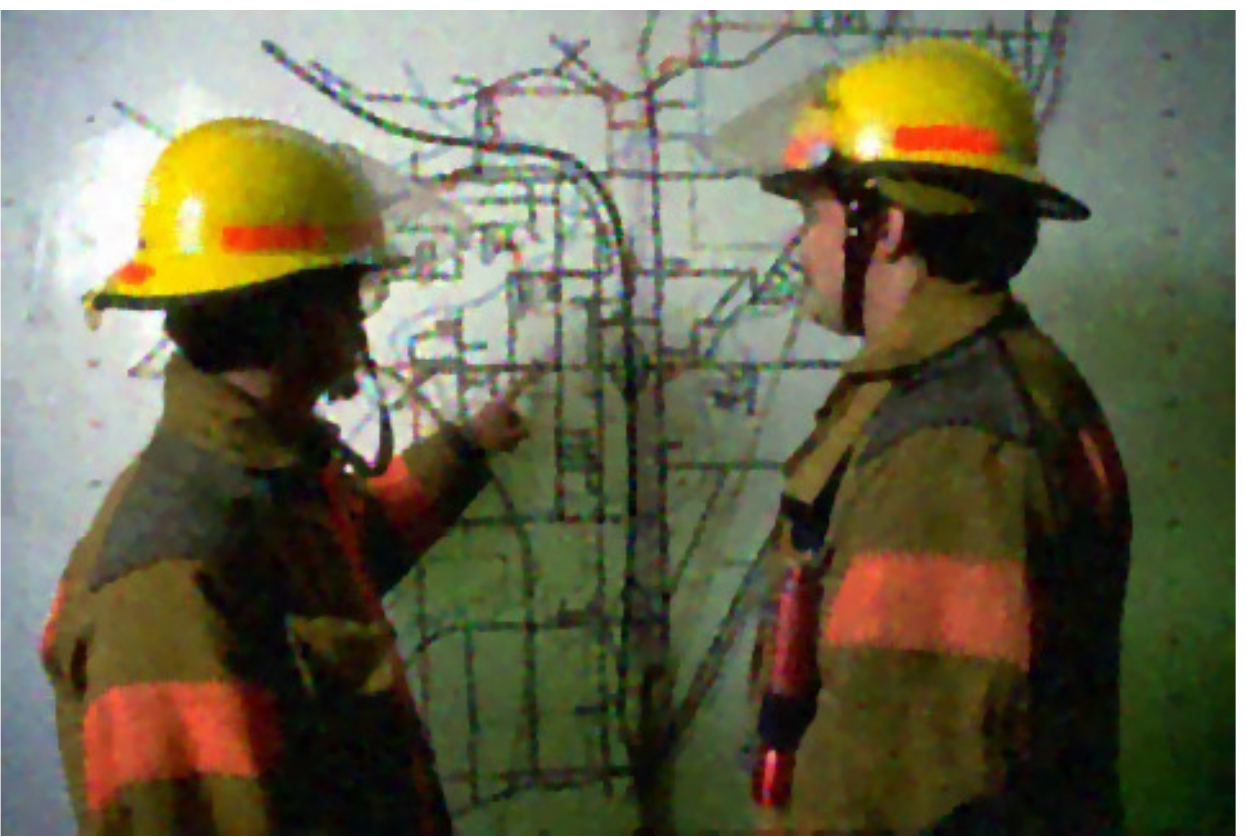}}

\subfloat[\textbf{$\bell_1$-ST-NLTV: 18.20 dB.}]{\includegraphics[width=0.3\textwidth,crop]{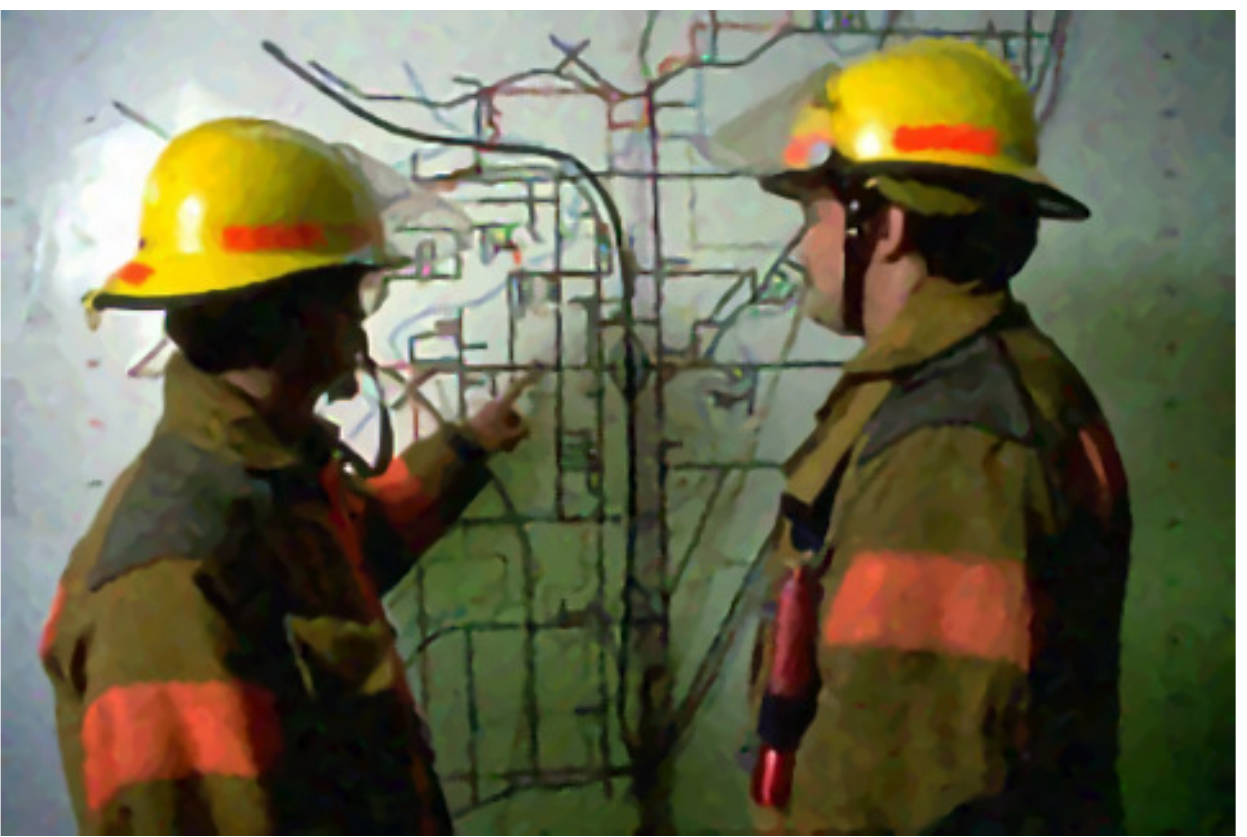}}
\hfill
\subfloat[$\bell_2$-ST-NLTV: 17.46 dB.]{\includegraphics[width=0.3\textwidth,crop]{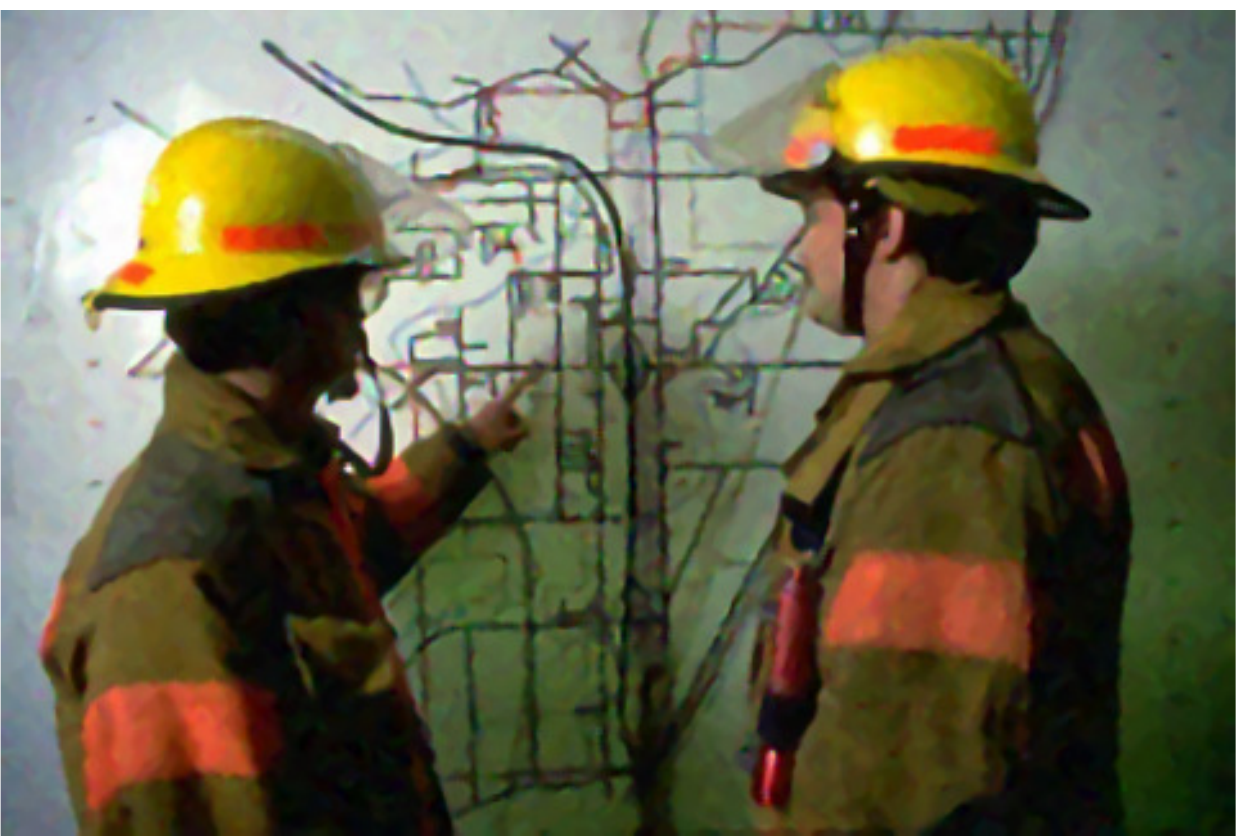}}
\hfill
\subfloat[$\bell_\infty$-ST-NLTV: 16.67 dB.]{\includegraphics[width=0.3\textwidth,crop]{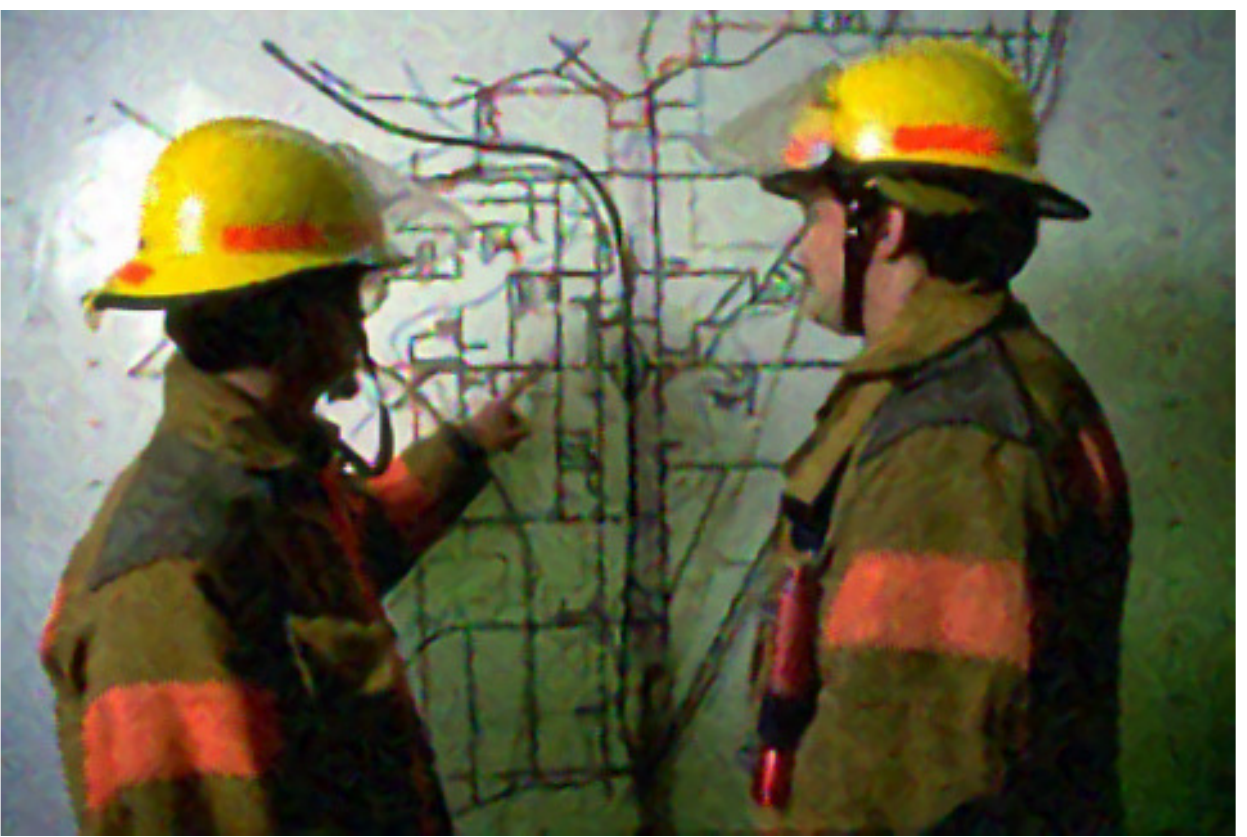}}

\caption{Visual comparison of a color image reconstructed with various regularization constraints. Degradation: additive zero-mean white Gaussian noise with std.\ deviation equal to $10$, uniform blur of size $3\times 3$, and $80\%$ of decimation ($N=154401$, $R=S=3$ and $K=30880$).}
\label{fig:firemen}
\end{figure*}

In the example of Fig.~\ref{fig:firemen}, we collect the images reconstructed by using $\bell_p$-CC-TV \cite{Attouche_H_2006_siam_variational_asb}, $\bell_p$-CC-NLTV \cite{Chierchia_G_2012_j-ieee-tsp_epigraphical_ppt}, $\bell_p$-ST-TV \cite{Bresson_X_2008_fastdual,Goldluecke_B_2012_natural_tv_gmt} and the proposed $\bell_p$-ST-NLTV for $p \in \{1,2,+\infty\}$. The results demonstrate the interest of considering non-local structure tensor measures, $\bell_1$-ST-NLTV  being the most effective regularization. Indeed, ST-NLTV regularization combines the advantages of both ST and NLTV, as one can see a better preservation of details and a reduction of color smearing.

\makeatletter
\define@key{Gin}{crop}[true]{%
    \edef\@tempa{{Gin}{trim=33mm 20mm 40mm 30mm,clip}}%
    \expandafter\setkeys\@tempa
}
\makeatother

\subsection{Imaging spectroscopy}
In this section, we compare $\bell_1$-ST-NLTV with implementations of two state-of-the-art methods in spectral imagery: Hyperspectral TV (H-TV) \cite{Yuan_2012_j-ieee-tgrs_hyper_denoising_TV} (see Section~\ref{sec:ST-TV}), and Multichannel NLTV (M-NLTV) \cite{Cheng_2014_j-tgrs_inpaint_images_MNLTV} (see Section~\ref{sec:ST-NLTV}). To this end, two scenarios are addressed by using the degradation model in \eqref{eq:cs_model}: a compressive-sensing scenario in which the measurement operator $(\textrm D_r)_{1\le r\le R}$ is a random decimation, and a restoration scenario in which $(\textrm D_r)_{1\le r\le R}$ is a decimated convolution. For reproducibility purposes, we use several publicly available multispectral and hyperspectral images.\footnote{https://engineering.purdue.edu/$\sim$biehl/MultiSpec/hyperspectral.html} The SNR index is used to give a quantitative assessment of the results obtained from the simulated experiments, reporting both the SNR computed over all the image and the average of SNR indices evaluated component-by-component (M-SNR).

In our experiments, the fidelity term related to the noise log-likelihood is $f = \|\textrm A\cdot-\textrm z\|^2_2$. Before degrading the original images, the pixels of each component are normalized in $[0, 255]$, hence the dynamic range constraint set $C$ imposes that the pixel values belong to $[0, 255]$. For the ST-NLTV constraints, we set $\tau_\ell\equiv 1$, $Q = 11$, $\widetilde{Q} = 5$, $\delta = 35$ and $\overline{M} = 14$.

\begin{figure*}
\centering
\subfloat[Component $r=81$.]{\includegraphics[width=0.3\textwidth]{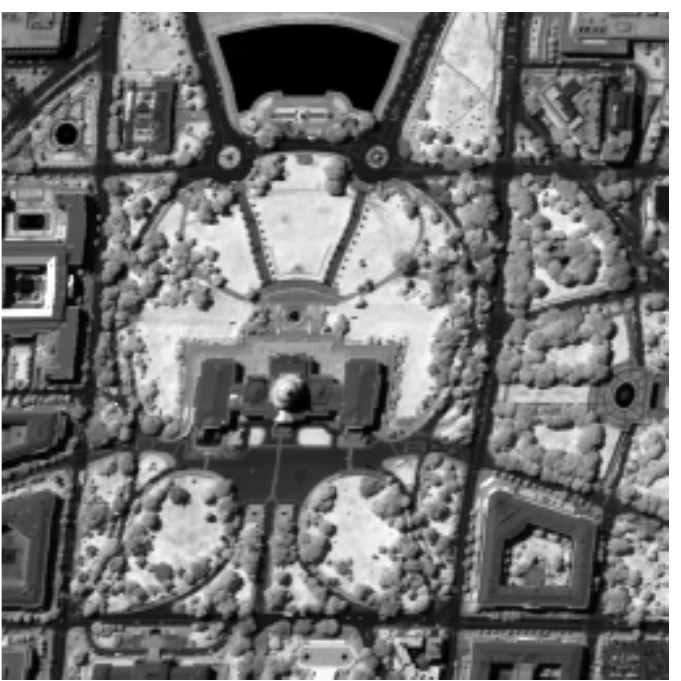}}
\hfill
\subfloat[H-TV: 11.78 dB.]{\includegraphics[width=0.3\textwidth]{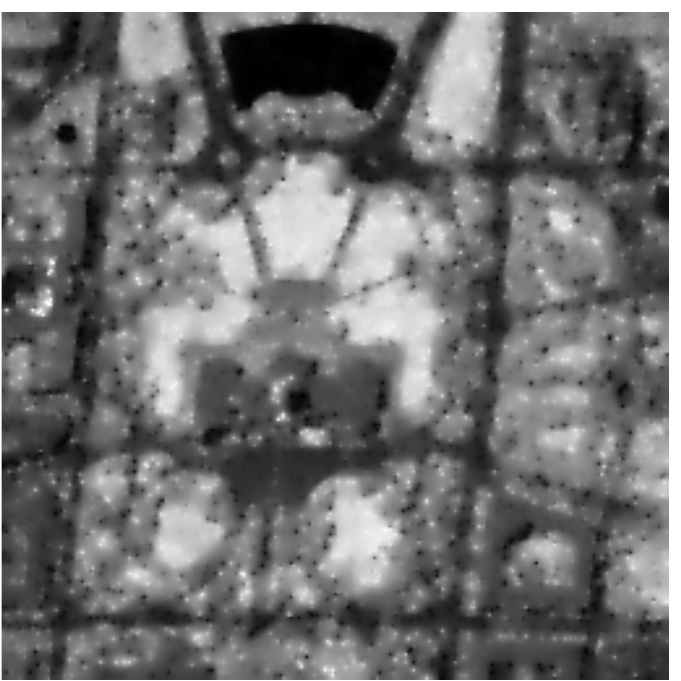}}
\hfill
\subfloat[$\bell_1$-ST-TV: 12.98 dB.]{\includegraphics[width=0.3\textwidth]{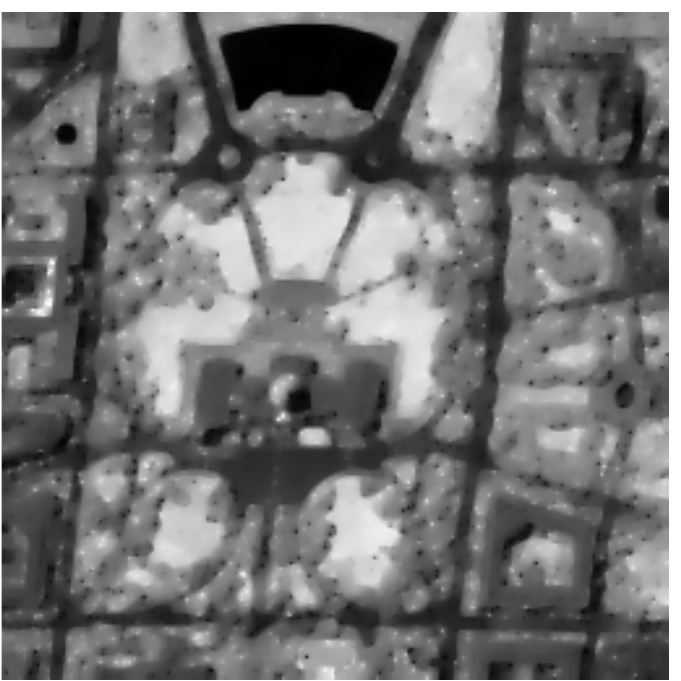}}

\subfloat[Noisy.]{\includegraphics[width=0.3\textwidth]{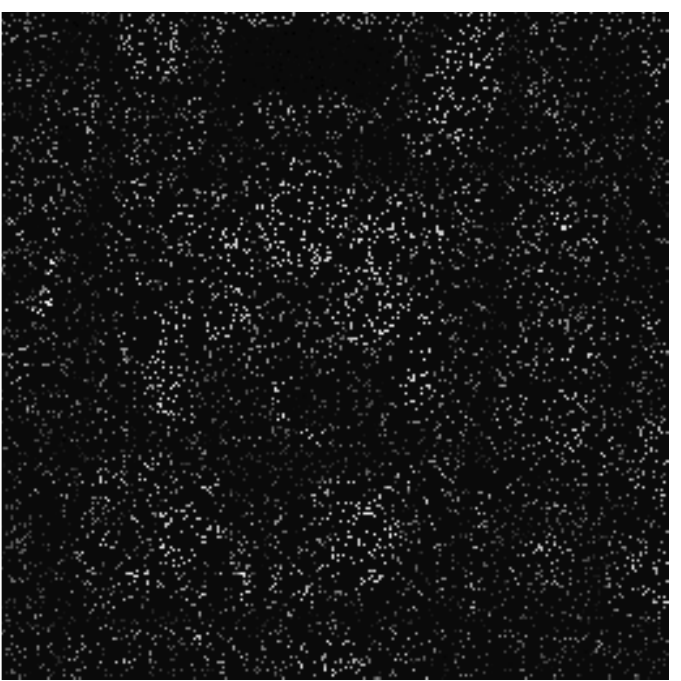}}
\hfill
\subfloat[M-NLTV: 12.76 dB.]{\includegraphics[width=0.3\textwidth]{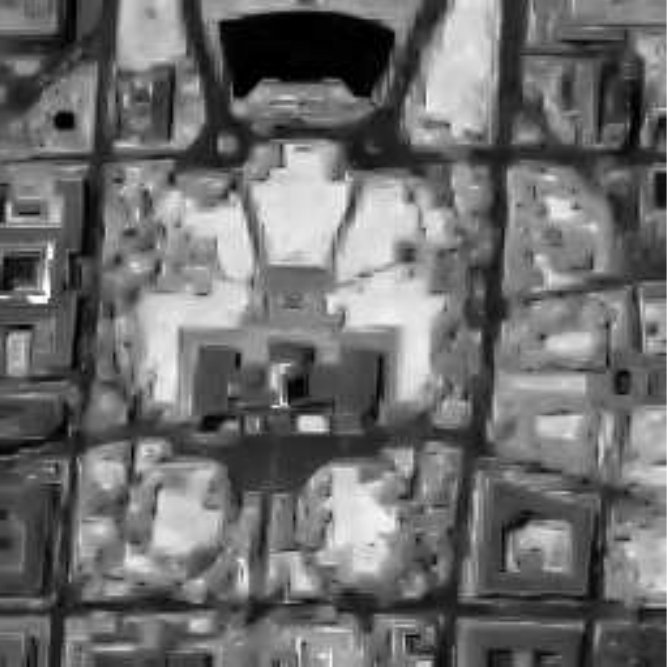}}
\hfill
\subfloat[\textbf{$\bell_1$-ST-NLTV: 14.36 dB}.]{\includegraphics[width=0.3\textwidth]{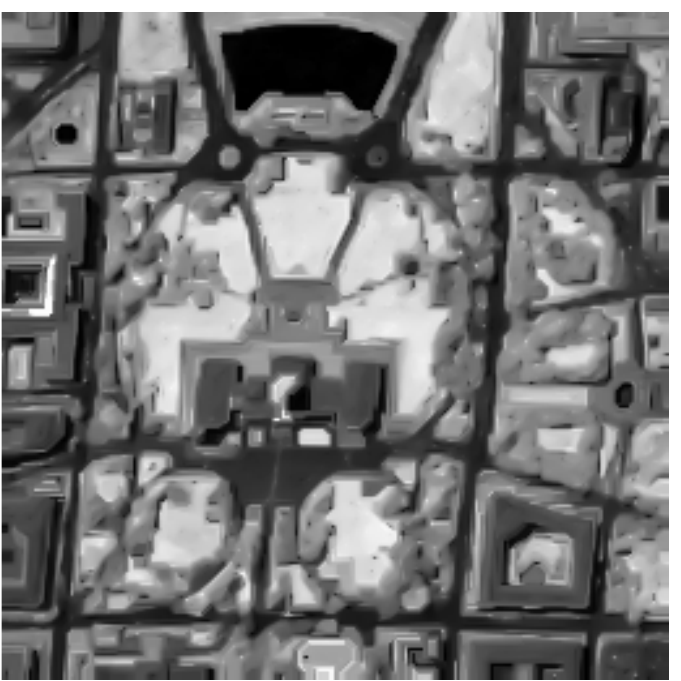}}

\caption{Visual comparison of the hyperspectral image \emph{hydice} reconstructed with H-TV \cite{Yuan_2012_j-ieee-tgrs_hyper_denoising_TV}, $\bell_1$-ST-TV, M-NLTV \cite{Cheng_2014_j-tgrs_inpaint_images_MNLTV} and $\bell_1$-ST-NLTV. Degradation: compressive sensing scenario involving an additive zero-mean white Gaussian noise with std.\ deviation $5$ and $90\%$ of decimation ($N=65536$, $R=191$, $K=6553$ and $S=191$).}
\label{fig:hydice_visual}
\end{figure*}

\begin{figure*}
\centering
\subfloat[SNR (dB) vs component index (image: hydice).]{\includegraphics[width=0.45\textwidth]{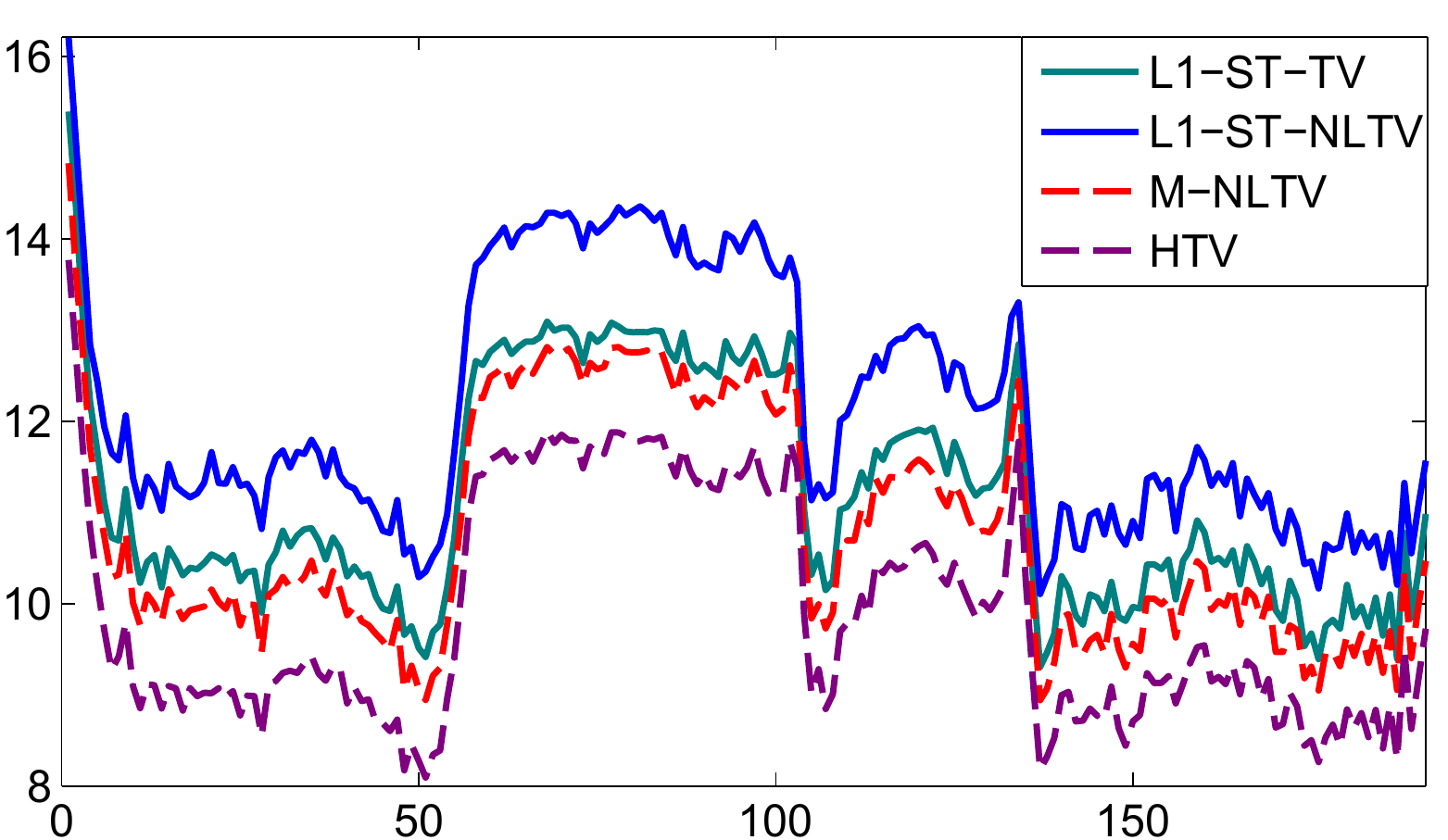}}
\hfill
\subfloat[SNR (dB) vs component index (image: indian pine).]{\includegraphics[width=0.45\textwidth]{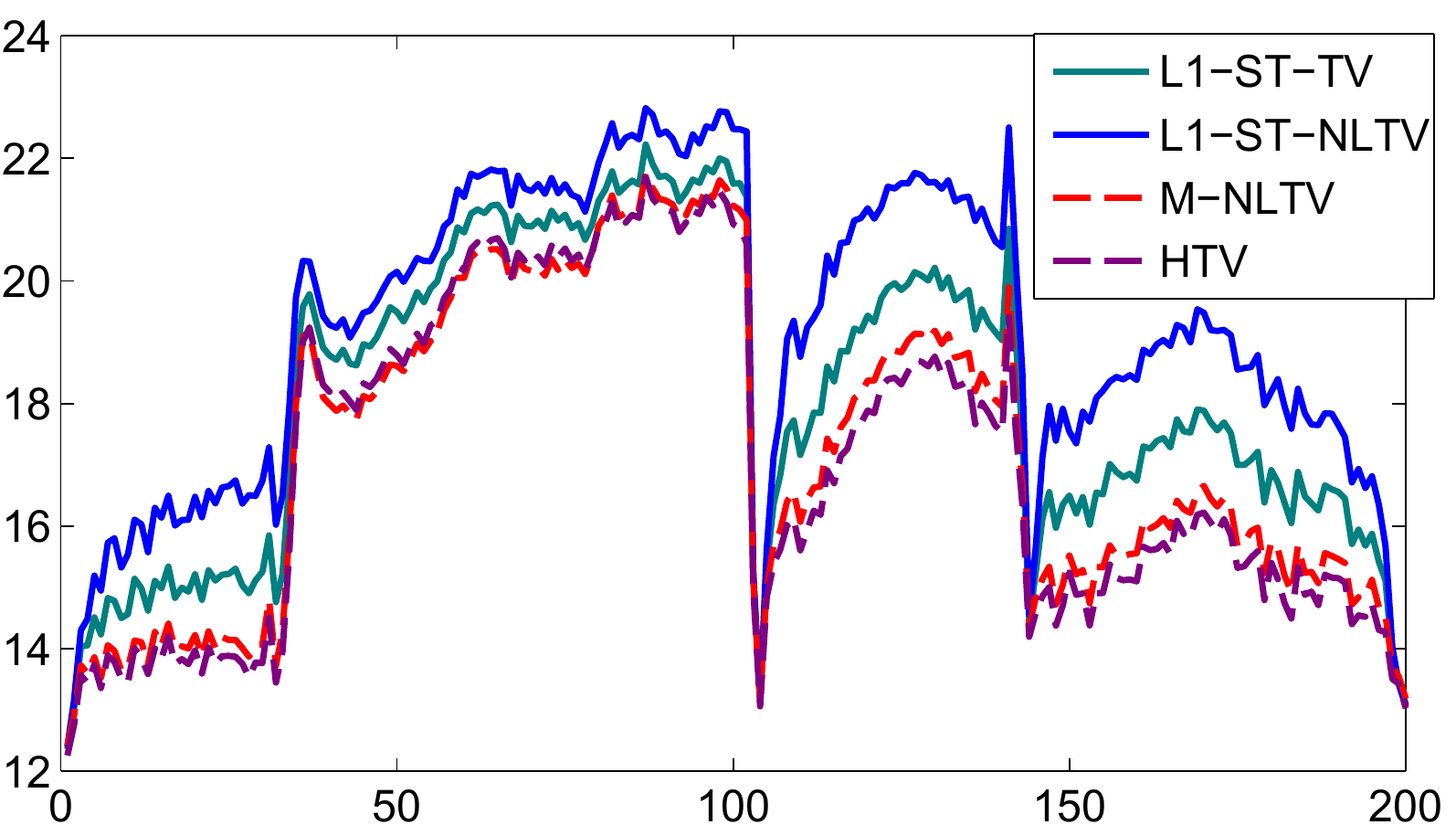}}
\caption{Quantitative comparison of two hyperspectral images reconstructed with H-TV \cite{Yuan_2012_j-ieee-tgrs_hyper_denoising_TV}, $\bell_1$-ST-TV, M-NLTV \cite{Cheng_2014_j-tgrs_inpaint_images_MNLTV} and $\bell_1$-ST-NLTV. Degradation: compressive sensing scenario involving an additive zero-mean white Gaussian noise with std.\ deviation $5$ and $90\%$ of decimation.}
\label{fig:hydice_plot}
\end{figure*}

Extensive tests have been carried out on several images of different sizes. The SNR and M-SNR indices obtained by using the proposed $\bell_1$-ST-TV and $\bell_1$-ST-NLTV regularization constraints are collected in Tables \ref{tab:multi1} and \ref{tab:multi2} for the two degradation scenarios mentioned above. In addition, a comparison is performed between our method and the H-TV and M-NLTV algorithms mentioned above (using an M+LFBF implementation). The hyper-parameter for each method (the bound $\eta$ for the ST constraint in our algorithm) was hand-tuned in order to achieve the best SNR values. The best results are highlighted in bold. Moreover, a component-by-component comparison of two hyperspectral images is made in \figurename~\ref{fig:hydice_plot}, while a visual comparison of a component from the image \emph{hydice} is displayed in \figurename~\ref{fig:hydice_visual}. 

The aforementioned results demonstrate the interest of combining the non-locality principle with structure-tensor smoothness measures. Indeed, $\bell_1$-ST-NLTV proves to be the most effective regularization with gains in SNR (up to 1.4~dB) with respect to M-NLTV, which in turn is comparable with $\bell_1$-ST-TV. The better performance of $\bell_1$-ST-NLTV seems to be related to its ability to better preserve edges and thin structures present in images, while preventing component smearing. 

\begin{table*}
\centering
\caption{SNR (dB) -- Mean SNR (dB) of reconstructed images (Degradation: std.\ deviation = $5$, decimation = $90\%$).}
\label{tab:multi1}
\begin{tabular}{llcccc}
\toprule
image & \multicolumn{1}{c}{size} & H-TV \cite{Yuan_2012_j-ieee-tgrs_hyper_denoising_TV} & $\bell_1$-ST-TV & M-NLTV \cite{Cheng_2014_j-tgrs_inpaint_images_MNLTV} & $\bell_1$-ST-NLTV \\
\midrule 
Hydice         & $256\times 256\times 191$ & 10.65 -- 09.87 & 11.93 -- 11.16 & 11.57 -- 10.76 & \textbf{12.98 -- 12.11} \\
Indian Pine    & $145\times 145\times 200$ & 17.31 -- 17.00 & 18.46 -- 18.24 & 17.62 -- 17.34 & \textbf{19.53 -- 19.49} \\
Little Coriver & $512\times 512\times 7$   & 17.81 -- 18.20 & 18.49 -- 18.83 & 18.46 -- 18.90 & \textbf{19.88 -- 20.18} \\
Mississippi    & $512\times 512\times 7$   & 18.27 -- 18.07 & 18.60 -- 18.37 & 18.94 -- 18.59 & \textbf{19.56 -- 19.28} \\
Montana        & $512\times 512\times 7$   & 22.49 -- 20.97 & 22.68 -- 21.15 & 22.85 -- 21.29 & \textbf{23.31 -- 21.76} \\
Rio            & $512\times 512\times 7$   & 16.48 -- 15.29 & 16.65 -- 15.48 & 16.82 -- 15.64 & \textbf{17.20 -- 16.05} \\
Paris          & $512\times 512\times 7$   & 14.85 -- 14.31 & 14.94 -- 14.39 & 15.05 -- 14.53 & \textbf{15.36 -- 14.82} \\
\bottomrule
\end{tabular}
\end{table*}

\begin{table*}
\centering
\caption{SNR (dB) -- mean SNR (dB) of restored images (Degradation: std.\ deviation = $5$, blur = $5\times 5$, decimation = $70\%$).}
\label{tab:multi2}
\begin{tabular}{llcccc}
\toprule
image & \multicolumn{1}{c}{size} & H-TV \cite{Yuan_2012_j-ieee-tgrs_hyper_denoising_TV} & $\bell_1$-ST-TV & M-NLTV \cite{Cheng_2014_j-tgrs_inpaint_images_MNLTV} & $\bell_1$-ST-NLTV \\
\midrule 
Hydice         & $256\times 256\times 191$ & 13.76 -- 12.90 & 14.30 -- 13.50 & 13.84 -- 12.98 & \textbf{14.84 -- 14.08}  \\
Indian Pine    & $145\times 145\times 200$ & 19.80 -- 19.65 & 20.22 -- 20.13 & 19.73 -- 19.57 & \textbf{20.43 -- 20.41}  \\
Little Coriver & $512\times 512\times 7$   & 21.35 -- 21.88 & 21.62 -- 22.01 & 21.31 -- 22.00 & \textbf{21.99 -- 22.49}  \\
Mississippi    & $512\times 512\times 7$   & 21.12 -- 20.29 & 21.21 -- 20.27 & 21.41 -- 20.52 & \textbf{21.65 -- 20.83}  \\
Montana        & $512\times 512\times 7$   & 24.80 -- 23.37 & 24.82 -- 23.31 & 24.96 -- 23.53 & \textbf{25.18 -- 23.72}  \\
Rio            & $512\times 512\times 7$   & 18.62 -- 17.50 & 18.57 -- 17.48 & 18.57 -- 17.60 & \textbf{18.87 -- 17.80}  \\
Paris          & $512\times 512\times 7$   & 16.68 -- 16.55 & 16.80 -- 16.53 & 16.73 -- 16.60 & \textbf{17.05 -- 16.81}  \\
\bottomrule
\end{tabular}
\end{table*}

\subsection{Comparison with SDMM}\label{sec:results:sdmm}
To complete our analysis, we compare the execution time of Algorithm~\ref{algo:epi} with respect to three alternative solutions:
\begin{itemize}
\item M+LFBF applied to Problem~\eqref{e:prob2} by computing the projection onto $D$ via the procedure  in \cite{VanDenBerg_E_2008_j-siam-sci-comp_pro_pfb};
\item SDMM applied to Problem~\eqref{eq:prob_sdmm} by computing the projection onto $D$ via the procedure  in \cite{VanDenBerg_E_2008_j-siam-sci-comp_pro_pfb} (Algorithm~\ref{algo:sdmm});
\item SDMM applied to Problem~\eqref{eq:prob_sdmm} after that the constraint $D$ is replaced by the constraints $E$ and $W$.
\end{itemize}
We would like to emphasize that all the above algorithms solve \emph{exactly} Problem~\eqref{e:prob2}, hence they produce equivalent results (i.e.\ they converge to the same solution). Our objective here is to empirically demonstrate that the epigraphical splitting technique and primal-dual proximal algorithms constitute a competitive choice for the problem at hand.

We present the results obtained with the image \emph{indian pine}, since a similar behaviour was observed for other images. The stopping criterion is set to $\norm{x^{[i+1]}-x^{[i]}} \le 10^{-5} \norm{x^{[i]}}$. We developed in MATLAB the basic structure of the aforementioned algorithms, while the most ``complex'' operations (such as the non-local gradient and projection computations) were implemented in C using mex files. In order to compute the projection onto $D$, we used the $\bell_{1}$-ball projector described in \cite[Algorithm 2]{VanDenBerg_E_2008_j-siam-sci-comp_pro_pfb},\footnote{The mex-file is available at \texttt{www.cs.ubc.ca/$\sim$mpf/spgl1}} as it avoids the expensive sorting operation (a review of several $\bell_{1}$-ball projectors can be found in \cite{Condat2014_fast_proj_L1}). Our codes were executed in Matlab R2011b with an Intel Xeon CPU at 2.80 GHz and 8 GB of RAM.

\begin{figure*}
\centering
\subfloat[$\bell_1$-ST-TV.]{\includegraphics[width=0.49\textwidth]{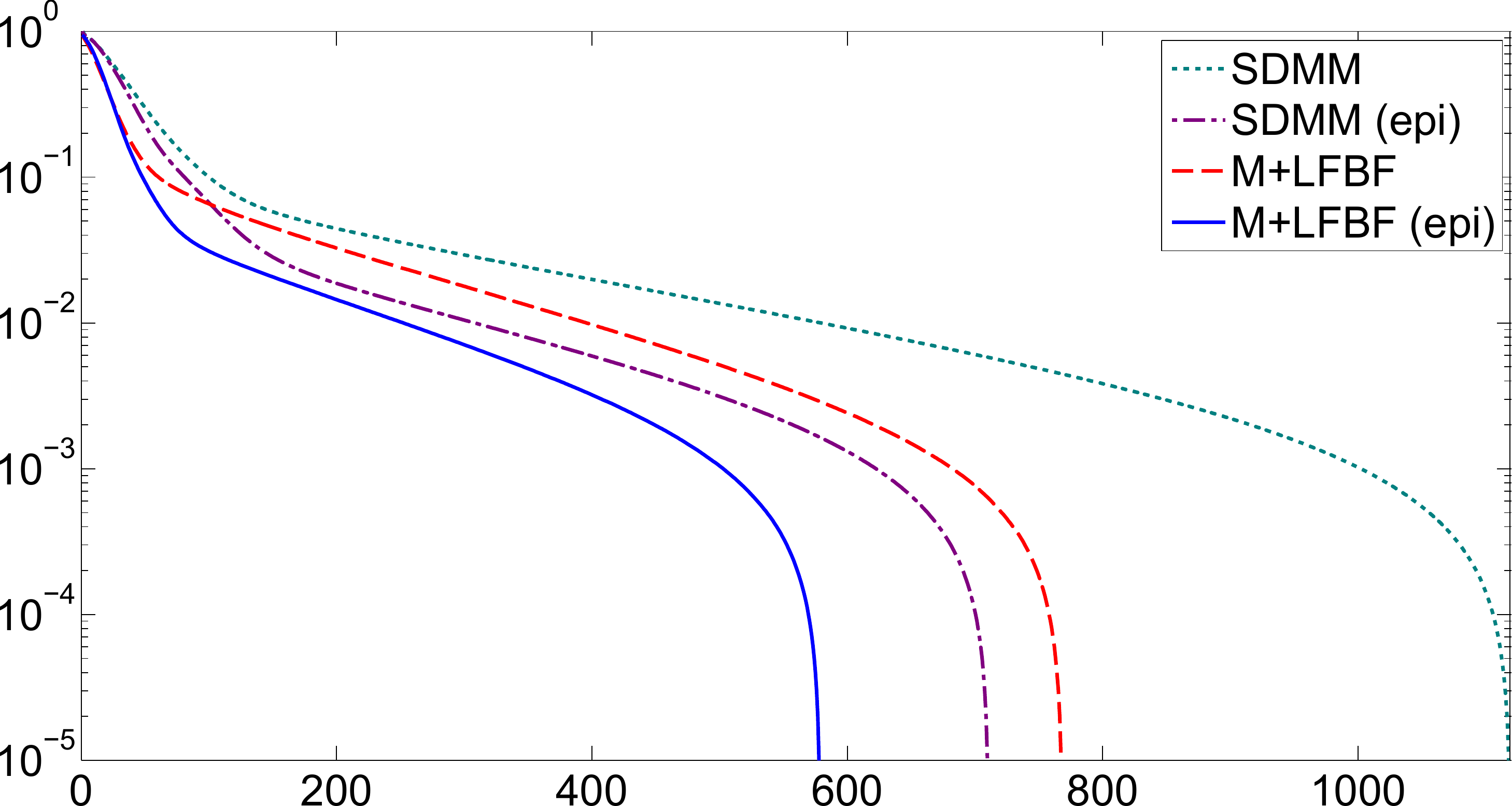}}
\hfill
\subfloat[$\bell_1$-ST-NLTV.]{\includegraphics[width=0.49\textwidth]{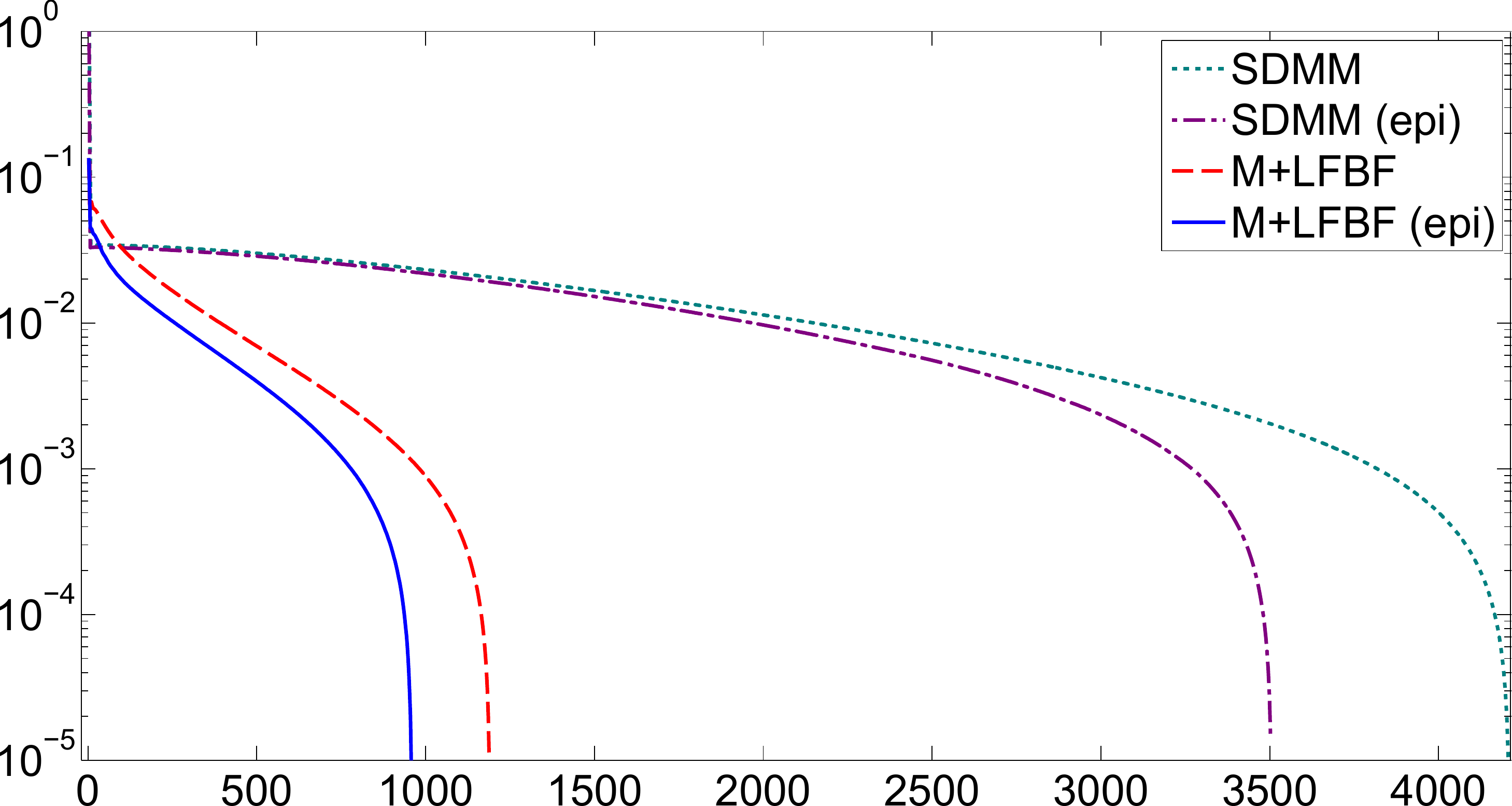}}
\caption{Comparison between epigraphical and direct methods: $\frac{\norm{x^{[i]}-x^{[\infty]}}}{\norm{x^{[\infty]}}}$ vs time (Degradation: std.\ deviation = $5$, decimation = $90\%$).}
\label{fig:exec_time}
\end{figure*}

Fig.~\ref{fig:exec_time} shows the relative error $\norm{x^{[i]}-x^{[\infty]}}/\norm{x^{[\infty]}}$ as a function of the computational time, where $x^{[\infty]}$ denotes the solution computed with a stopping criterion of $10^{-5}$. These plots indicate that the epigraphical approach yields a faster convergence than the direct one for both SDMM and M+LFBF, the latter being much faster than the former. This can be explained by the computational cost of the subiterations required by the direct projection onto the $\bell_{1}$-ball. Note that these conclusions extend to all images in the dataset.

The results in Fig.~\ref{fig:exec_time} refer to the constraint bound $\eta$ that achieves the best SNR indices. In practice, the optimal bound may not be known precisely, although a reasonable estimate may be available for certain classes of images based on statistics of databases \cite{Combettes_PL_2004_tip_TV_irstavc}. While it is out of the scope of this paper to investigate an optimal strategy to set this bound, it is important to evaluate the impact of its choice on our method  performance. In Tables~\ref{tab:st_tv}~and~\ref{tab:st_nltv}, we compare the epigraphical approach with the direct computation of the projections (via standard iterative solutions) for different choices of $\eta$. For better readability, the values of $\eta$ are expressed as a multiplicative factor of the ST-TV and ST-NLTV semi-norms of the original image. The execution times indicate that the epigraphical approach yields a faster convergence than the direct approach for SDMM and M+LFBF. Moreover, the numerical results show that errors within $\pm 5\%$ from the optimal value for $\eta$ lead to SNR variations within $1.2\%$. We refer to \cite{Chierchia_G_2012_j-ieee-tsp_epigraphical_ppt} for an extensive comparison between the epigraphical and direct approaches.

\begin{table*}
  \centering%
  \caption{Results for the $\bell_1$-ST-TV constraint and some values of $\eta$. Degradation: std.\ deviation = $5$, decimation = $90\%$.\hspace{\textwidth}{\scriptsize(``speed up'' is the ratio between ``direct'' and ``epigraphical'' times)}}
  {\scriptsize
  \begin{tabular}{+c@{\quad}^c @{\quad} ^c@{\;}^c ^c@{\;}^c @{}^c @{\;}^c@{}^c @{\qquad} ^c@{\;}^c ^c@{\;}^c @{}^c @{\;} ^c}
    \toprule
    \multirow{4}{*}{$\eta$} & \multirow{4}{*}{SNR (dB) -- M-SNR (dB)} & \multicolumn{6}{c}{SDMM} && \multicolumn{6}{c}{M+LFBF} \\
    \cmidrule{3-8}\cmidrule(r){10-15}
    & & \multicolumn{2}{c}{direct} & \multicolumn{2}{c}{epigraphical} && \multirow{2}{*}{speed up} && \multicolumn{2}{c}{direct} & \multicolumn{2}{c}{epigraphical} && \multirow{2}{*}{speed up}\\
    \cmidrule(r){3-4}\cmidrule(lr){5-6}\cmidrule(r){10-11}\cmidrule(lr){12-13}
                            &                           & \# iter. & {sec.}    & \# iter. & {sec.}      && && \# iter. & {sec.}    & \# iter. & {sec.}      &&\\
    \midrule
		0.35 & 18.41 -- 18.19 & 547 &  767.51 & 471 & 466.80 && 1.64 && 466 & 471.95 & 389 & 339.24 && 1.39 \\
\brow	0.40 & 18.46 -- 18.24 & 838 & 1066.24 & 698 & 701.03 && 1.52 && 733 & 735.36 & 621 & 558.37 && 1.32 \\
		0.45 & 18.26 -- 18.02 &1000 & 1353.13 &1000 & 990.76 && 1.37 &&1000 &1018.58 &1000 & 902.00 && 1.13 \\
    \bottomrule
  \end{tabular}
  }
  \label{tab:st_tv}
\end{table*}

\begin{table*}
	\centering%
	\caption{Results for the $\bell_1$-ST-NLTV constraint and some values of $\eta$. Degradation: std.\ deviation = $5$, decimation = $90\%$.\hspace{\textwidth}{\scriptsize(``speed up'' is the ratio between ``direct'' and ``epigraphical'' times)}}
	{\scriptsize
	\begin{tabular}{+c@{\quad}^c @{\quad} ^c@{\;}^c ^c@{\;}^c @{}^c @{\;}^c@{}^c @{\qquad} ^c@{\;}^c ^c@{\;}^c @{}^c @{\;} ^c}
		\toprule
		\multirow{4}{*}{$\eta$} & \multirow{4}{*}{SNR (dB) -- M-SNR (dB)} & \multicolumn{6}{c}{SDMM} && \multicolumn{6}{c}{M+LFBF} \\
		\cmidrule{3-8}\cmidrule(r){10-15}
		& & \multicolumn{2}{c}{direct} & \multicolumn{2}{c}{epigraphical} && \multirow{2}{*}{speed up} && \multicolumn{2}{c}{direct} & \multicolumn{2}{c}{epigraphical} && \multirow{2}{*}{speed up}\\
		\cmidrule(r){3-4}\cmidrule(lr){5-6}\cmidrule(r){10-11}\cmidrule(lr){12-13}
		                   &                           & \# iter. & {sec.}    & \# iter. & {sec.}      && && \# iter. & {sec.}    & \# iter. & {sec.}      &&\\
		\midrule

		\multicolumn{15}{c}{\textsl{Neighbourhood size: $Q = 3$}}\\
		0.25 & 19.15 -- 19.05 & 1000 & 4384.38 & 1000 & 3583.57 && 1.23 && 190 &  494.77 & 190 & 448.22 && 1.11 \\
		0.30 & 19.39 -- 19.32 & 1000 & 4414.94 & 1000 & 3417.18 && 1.29 && 243 &  649.31 & 236 & 534.50 && 1.21 \\
		0.35 & 19.36 -- 19.28 &  875 & 4175.52 & 1000 & 3482.80 && 1.20 && 319 &  839.86 & 308 & 726.50 && 1.16 \\

		\multicolumn{15}{c}{\textsl{Neighbourhood size: $Q = 5$}}\\
		0.25 & 19.43 -- 19.38 & 1000 & 14412.86 & 1000 & 10167.34 && 1.42 && 216 &  977.95 & 212 & 871.80  && 1.12 \\
\brow	0.30 & 19.55 -- 19.51 & 1000 & 14338.36 & 1000 & 10174.68 && 1.41 && 275 & 1257.71 & 268 & 1143.35 && 1.10 \\
		0.35 & 19.53 -- 19.49 & 1000 & 14365.92 & 1000 & 10356.73 && 1.39 && 358 & 1631.17 & 347 & 1424.72 && 1.14 \\
		\bottomrule
	\end{tabular}
	}
	\label{tab:st_nltv}
\end{table*}

\section{Conclusions}\label{sec:conclusion}
We have proposed a new regularization for multicomponent images that is a combination of \textit{non-local total variation} and \textit{structure tensor}. The resulting image recovery problem has been formulated as a constrained convex optimization problem and solved through a novel epigraphical projection method using primal-dual proximal algorithms. The obtained results demonstrate the better performance of structure tensor and non-local gradients over a number of multispectral and hyperspectral images, although it would be interesting to consider other applications, such as the recovery of video signals or volumetric images. Our results also show that the nuclear norm has to be preferred over the Frobenius norm for hyperspectral image recovery problems. Furthermore, the experimental part indicates that the epigraphical method converges faster than the approach based on the direct computation of the projections via standard iterative solutions. In both cases, the proposed algorithm turns out to be faster than solutions based on the Alternating Direction Method of Multipliers, suggesting that primal-dual proximal algorithms constitute a good choice in practice to deal with multicomponent image recovery problems.

\bibliographystyle{IEEEbib}
\bibliography{abbr,biblio}

\end{document}